\documentclass{article} 
\usepackage{nips14submit_e}
\usepackage{times}
\usepackage[sort&compress,numbers]{natbib}
\usepackage{amsmath}
\usepackage{graphicx}
\usepackage{url}

\usepackage{algorithm}
\usepackage{algorithmic}

\usepackage{etoolbox}
\newcommand{\algorithmicdoinparallel}{\textbf{do in parallel}}
\makeatletter
\AtBeginEnvironment{algorithmic}{%
  \newcommand{\FORALLP}[2][default]{\textbf{for all machine} #2\ %
    \algorithmicdoinparallel\ALC@com{#1}\begin{ALC@for}}%
}

\usepackage{tabularx}
\usepackage{mdwlist}

\usepackage{tikz}
\usetikzlibrary{positioning,patterns,shapes}
\usepackage{pgfplots}
\usepgfplotslibrary{external}
\usepackage{multirow}
\usepackage{rotating}
\input{./Definitions}

\newcommand{\trank}{\text{rank}}

\title{Ranking via Robust Binary Classification}

\author{
Hyokun Yun \\
Department of Statistics\\
Purdue University\\
West Lafayette, IN 47907 \\
\texttt{yun3@purdue.edu} \\
\And
Parameswaran Raman \\
Department of Computer Science \\
Purdue University \\
West Lafayette, IN 47907 \\
\texttt{params@purdue.edu} \\
\And
S.~V.~N.~Vishwanathan\\
Departments of Statistics and Computer Science\\
Purdue University \\
West Lafayette, IN 47907 \\
\texttt{vishy@stat.purdue.edu} \\
}

\nipsfinalcopy 

\begin{document}

\maketitle

\begin{abstract}
  We propose RoBiRank, a ranking algorithm that is motivated by
  observing a close connection between evaluation metrics for learning
  to rank and loss functions for robust classification.  It shows
  competitive performance on standard benchmark datasets against a
  number of other representative algorithms in the literature.  We also
  discuss extensions of RoBiRank to large scale problems where explicit
  feature vectors and scores are not given. We show that RoBiRank can be
  efficiently parallelized across a large number of machines; for a task
  that requires $386,133 \times 49,824,519$ pairwise interactions
  between items to be ranked, RoBiRank finds solutions that are of
  dramatically higher quality than that can be found by a
  state-of-the-art competitor algorithm, given the same amount of
  wall-clock time for computation.
\end{abstract}

\section{Introduction}
\label{sec:Introduction}

Learning to rank is a problem of ordering a set of items according to
their relevances to a given context \citep{ChaCha11}.  While a number of
approaches have been proposed in the literature, in this paper we
provide a new perspective by showing a close connection between ranking
and a seemingly unrelated topic in machine learning, namely, robust
binary classification.

In robust classification \citep{Huber81}, we are asked to learn a
classifier in the presence of outliers.  Standard models for
classification such as Support Vector Machines (SVMs) and logistic
regression do not perform well in this setting, since the convexity of
their loss functions does not let them give up their performance on any
of the data points \citep{LonSer10}; for a classification model to be
robust to outliers, it has to be capable of sacrificing its performance
on some of the data points.  We observe that this requirement is very
similar to what standard metrics for ranking try to evaluate.
Discounted Cumulative Gain (DCG) \citep{ManRagSch08} and its normalized
version NDCG, popular metrics for learning to rank, strongly emphasize
the performance of a ranking algorithm at the top of the list;
therefore, a good ranking algorithm in terms of these metrics has to be
able to give up its performance at the bottom of the list if that can
improve its performance at the top.

In fact, we will show that DCG and NDCG can indeed be written as a
natural generalization of robust loss functions for binary
classification.  Based on this observation we formulate RoBiRank, a
novel model for ranking, which maximizes the lower bound of (N)DCG.
Although the non-convexity seems unavoidable for the bound to be tight
\citep{ChaDoTeoLeSmo08}, our bound is based on the class of robust loss
functions that are found to be empirically easier to optimize
\citep{Ding13}.  Indeed, our experimental results suggest that RoBiRank
reliably converges to a solution that is competitive as compared to
other representative algorithms even though its objective function is
non-convex.

While standard deterministic optimization algorithms such as L-BFGS
\citep{NocWri06} can be used to estimate parameters of RoBiRank, to
apply the model to large-scale datasets a more efficient parameter
estimation algorithm is necessary.  This is of particular interest in
the context of latent collaborative retrieval \citep{WesWanWeiBer12};
unlike standard ranking task, here the number of items to
rank is very large and explicit feature vectors and scores are not
given.

Therefore, we develop an efficient parallel stochastic optimization
algorithm for this problem.  It has two very attractive
characteristics: First, the time complexity of each stochastic update
is independent of the size of the dataset.  Also, when the algorithm
is distributed across multiple number of machines, no interaction
between machines is required during most part of the execution;
therefore, the algorithm enjoys near linear scaling.  This is a
significant advantage over serial algorithms, since it is very easy to
deploy a large number of machines nowadays thanks to the popularity of
cloud computing services, e.g.\ Amazon Web Services.

We apply our algorithm to latent collaborative retrieval task on Million
Song Dataset \citep{BerEllWhiLam13} which consists of 1,129,318 users,
386,133 songs, and 49,824,519 records; for this task, a ranking
algorithm has to optimize an objective function that consists of
$386,133 \times 49,824,519$ number of pairwise interactions.  With the
same amount of wall-clock time given to each algorithm, RoBiRank
leverages parallel computing to outperform the state-of-the-art with a
100\% lift on the evaluation metric.

\section{Robust Binary Classification}
\label{sec:binary_loss}


Suppose we are given training data which consists of $n$ data points
$(x_1, y_1), (x_2, y_2), \ldots, (x_n, y_n)$, where each $x_i \in \RR^d$
is a $d$-dimensional feature vector and $y_i \in \cbr{-1, +1}$ is a
label associated with it.  A linear model attempts to learn a
$d$-dimensional parameter $\omega$, and for a given feature vector $x$
it predicts label $+1$ if $\inner{x}{\omega} \geq 0$ and $-1$ otherwise.
Here $\inner{\cdot}{\cdot}$ denotes the Euclidean dot product between
two vectors.  The quality of $\omega$ can be measured by the number of
mistakes it makes:
$L(\omega) := \sum_{i=1}^n I (y_i \cdot \inner{x_i}{\omega} < 0)$.
The indicator function $I(\cdot < 0)$ is called the 0-1 loss function,
because it has a value of 1 if the decision rule makes a mistake, and 0
otherwise.  Unfortunately, since the 0-1 loss is a discrete function its
minimization is difficult~\citep{FelGurRagWu12}.  The most popular
solution to this problem in machine learning is to upper bound the 0-1
loss by an easy to optimize function \citep{JorBarMcA06}.  For example,
logistic regression uses the logistic loss function
$\sigma_0(t) := \log_2(1 + 2^{-t})$,
to come up with a continuous and convex objective function
\begin{align}
  \overline{L}(\omega) := \sum_{i=1}^n \sigma_0(y_i \cdot
  \inner{x_i}{\omega}),
  \label{eq:logistic_obj}
\end{align}
which upper bounds $L(\omega)$.  It is clear that for each $i$,
$\sigma_0(y_i \cdot \inner{x_i}{\omega})$ is a convex function in
$\omega$; therefore, $\overline{L}(\omega)$, a sum of convex
functions, is also a convex function which is relatively easier to
optimize \citep{BoyVan04}.  Support Vector Machines (SVMs) on the
other hand can be recovered by using the hinge loss to upper bound the
0-1 loss.

However, convex upper bounds such as $\overline{L}(\omega)$ are known to
be sensitive to outliers \citep{LonSer10}.  The basic intuition here is
that when $y_i \cdot \inner{x_i}{\omega}$ is a very large negative
number for some data point $i$, $\sigma(y_i \cdot \inner{x_i}{\omega})$
is also very large, and therefore the optimal solution of
\eqref{eq:logistic_obj} will try to decrease the loss on such outliers
at the expense of its performance on ``normal'' data points.

In order to construct robust loss functions, consider the following two
transformation functions:
\begin{align}
  \rho_1(t) := \log_2(t + 1), \;\; \rho_2(t) := 1 - \frac 1 {\log_2(t +
    2)},
  \label{eq:transforms}
\end{align}
which, in turn, can be used to define the following loss functions:
\begin{align}
  \sigma_1(t) &:= \rho_1(\sigma_0(t)), \;\; \sigma_2(t) :=
  \rho_2(\sigma_0(t)). \label{robust_losses}
\end{align}
One can see that $\sigma_1(t) \rightarrow \infty$ as $t \rightarrow
-\infty$, but at a much slower rate than $\sigma_0(t)$ does; its
derivative $\sigma'_1(t) \rightarrow 0$ as $t \rightarrow -\infty$.
Therefore, $\sigma_1(\cdot)$ does not grow as rapidly as $\sigma_0(t)$
on hard-to-classify data points.  Such loss functions are called Type-I
robust loss functions by \citet{Ding13}, who also showed that they enjoy
statistical robustness properties.  $\sigma_2(t)$ behaves even better:
$\sigma_2(t)$ converges to a constant as $t \rightarrow -\infty$, and
therefore ``gives up'' on hard to classify data points.  Such loss
functions are called Type-II loss functions, and they also enjoy
statistical robustness properties \citep{Ding13}.

In terms of computation, of course, $\sigma_1(\cdot)$ and
$\sigma_2(\cdot)$ are not convex, and therefore the objective function
based on such loss functions is more difficult to optimize.  However, it
has been observed in \citet{Ding13} that models based on optimization of
Type-I functions are often empirically much more successful than those
which optimize Type-II functions.  Furthermore, the solutions of Type-I
optimization are more stable to the choice of parameter
initialization. Intuitively, this is because Type-II functions asymptote
to a constant, reducing the gradient to almost zero in a large
fraction of the parameter space; therefore, it is difficult for a
gradient-based algorithm to determine which direction to pursue. See
\citet{Ding13} for more details. 

\section{Ranking Model via Robust Binary Classification}
\label{sec:RankingLoss}



Let $\Xcal = \cbr{x_1, x_2, \ldots, x_n }$ be a set of contexts, and
$\Ycal = \cbr{y_1, y_2, \ldots, y_m }$ be a set of items to be ranked.
For example, in movie recommender systems $\Xcal$ is the set of users
and $\Ycal$ is the set of movies.  In some problem settings, only a
subset of $\Ycal$ is relevant to a given context $x \in \Xcal$;
e.g. in document retrieval systems, only a subset of documents is
relevant to a query.  Therefore, we define $\Ycal_x \subset \Ycal$ to
be a set of items relevant to context $x$.  Observed data can be
described by a set $W := \cbr{ W_{xy} }_{x \in \Xcal, y \in \Ycal_x}$
where $W_{xy}$ is a real-valued score given to item $y$ in context
$x$.

We adopt a standard problem setting used in the literature of learning
to rank.  For each context $x$ and an item $y \in \Ycal_x$, we aim to
learn a scoring function $f(x,y): \Xcal \times \Ycal_x \rightarrow \RR$
that induces a ranking on the item set $\Ycal_x$; the higher the score,
the more important the associated item is in the given context.  To
learn such a function, we first extract joint features of $x$ and $y$,
which will be denoted by $\phi(x,y)$.  Then, we parametrize $f(\cdot,
\cdot)$ using a parameter $\omega$, which yields the linear model
$f_\omega(x,y) := \inner{\phi(x,y)}{\omega}$, where, as before,
$\inner{\cdot}{\cdot}$ denotes the Euclidean dot product between two
vectors.  $\omega$ induces a ranking on the set of items $\Ycal_x$; we
define $\trank_\omega(x,y)$ to be the rank of item $y$ in a given
context $x$ induced by $\omega$.  Observe that $\trank_\omega(x,y)$ can
also be written as a sum of 0-1 loss functions (see e.g.\
\citet{UsuBufGal09}):
\begin{align}
  \trank_\omega(x,y) = \mkern-18mu \sum_{y' \in \Ycal_x, y' \neq y}
  \mkern-18mu I\rbr{f_\omega(x,y) - f_\omega(x,y') < 0}.
  \label{eq:rank_as_sum}
\end{align}

\subsection{Basic Model}
\label{ssec:basic_model}

If an item $y$ is very relevant in context $x$, a good parameter
$\omega$ should position $y$ at the top of the list; in other words,
$\trank_\omega(x,y)$ has to be small, which motivates the following
objective function:
\begin{align}
  L(\omega) := \sum_{x \in \Xcal} c_x \sum_{y \in \Ycal_x} v(W_{xy})
  \cdot \trank_\omega(x,y),
  \label{eq:basic_loss}
\end{align}
where $c_x$ is an weighting factor for each context $x$, and $v(\cdot):
\RR^+ \rightarrow \RR^+$ quantifies the relevance level of $y$ on $x$.
Note that $\cbr{c_x}$ and $v(W_{xy})$ can be chosen to reflect the
metric the model is going to be evaluated on (this will be discussed in
Section~\ref{ssec:ndcg}).  Note that \eqref{eq:basic_loss} can be
rewritten using \eqref{eq:rank_as_sum} as a sum of indicator functions.
Following the strategy in Section~\ref{sec:binary_loss}, one can form an
upper bound of \eqref{eq:basic_loss} by bounding each 0-1 loss function
by a logistic loss function:
\begin{align}
  \overline{L}(\omega) := \sum_{x \in \Xcal} &c_x \sum_{y \in \Ycal_x}
  v\rbr{W_{xy}} \cdot \sum_{y' \in \Ycal_x, y' \neq y}
  \sigma_0\rbr{f_\omega(x,y) - f_\omega(x,y')}.
  \label{eq:bound_basic_loss}
\end{align}
Just like \eqref{eq:logistic_obj}, \eqref{eq:bound_basic_loss} is
convex in $\omega$ and hence easy to minimize.  


\subsection{DCG}
\label{ssec:ndcg}

Although \eqref{eq:bound_basic_loss} enjoys convexity, it may not be a
good objective function for ranking.  This is because in most
applications of learning to rank, it is more important to do well at the
top of the list than at the bottom, as users typically pay attention
only to the top few items.  Therefore, it is desirable to \emph{give up}
performance on the lower part of the list in order to gain quality at
the top.  This intuition is similar to that of robust classification in
Section~\ref{sec:binary_loss}; a stronger connection will be shown
below.

Discounted Cumulative Gain (DCG)~\citep{ManRagSch08} is one of the most
popular metrics for ranking.  For each context $x \in \Xcal$, it is
defined as:
\begin{align}
  \text{DCG}(\omega) := c_x 
  \sum_{y \in \Ycal_x}
  \frac{v\rbr{W_{xy}}}{\log_2\rbr{\trank_\omega(x,y) + 2}},
  \label{eq:gen_metric}
\end{align}
where $v(t) = 2^t - 1$ and $c_x = 1$. Since $1/\log(t + 2)$ decreases
quickly and then asymptotes to a constant as $t$ increases, this metric
emphasizes the quality of the ranking at the top of the list.
Normalized DCG (NDCG) simply normalizes the metric to bound it between 0
and 1 by calculating the maximum achievable DCG value $m_x$ and dividing
by it \citep{ManRagSch08}.

\subsection{RoBiRank}

Now we formulate RoBiRank, which optimizes the lower bound of metrics
for ranking in form \eqref{eq:gen_metric}.  Observe
that $\max_{\omega} \text{DCG} (\omega)$ can be rewritten as 
\begin{align}
  \min_\omega \sum_{x \in \Xcal} c_x \sum_{y \in \Ycal_x} v\rbr{W_{xy}}
  \cdot \cbr{ 1 - \frac{1}{\log_2\rbr{\trank_\omega(x,y) + 2}} }
  \label{eq:to_opt}.
\end{align}
Using \eqref{eq:rank_as_sum} and the definition of the transformation
function $\rho_2(\cdot)$ in \eqref{eq:transforms}, we can rewrite
the objective function in \eqref{eq:to_opt} as:
\begin{align}
  L_2(\omega) := \sum_{x \in \Xcal} c_x \sum_{y \in \Ycal_x}
  v\rbr{W_{xy}} \cdot \rho_2 \rbr{ \sum_{y' \in \Ycal_x, y' \neq y}
    I\rbr{f_\omega(x,y) - f_\omega(x,y') < 0}}.
  \label{eq:rankloss_2}
\end{align}

Since $\rho_2(\cdot)$ is a monotonically increasing function, we can
bound \eqref{eq:rankloss_2} with a continuous function by bounding
each indicator function using the logistic loss:
\begin{align}
  \overline{L}_2(\omega) := \sum_{x \in \Xcal} c_x \sum_{y \in \Ycal_x}
  v\rbr{W_{xy}} \cdot \rho_2 \rbr{ \sum_{y' \in \Ycal_x, y' \neq y}
    \sigma_0\rbr{f_\omega(x,y) - f_\omega(x,y')}}.
  \label{eq:rankloss_2_bd}
\end{align}
This is reminiscent of the basic model in \eqref{eq:bound_basic_loss};
as we applied the transformation function $\rho_2(\cdot)$ on the
logistic loss function $\sigma_0(\cdot)$ to construct the robust loss
function $\sigma_2(\cdot)$ in \eqref{robust_losses}, we are again
applying the same transformation on \eqref{eq:bound_basic_loss} to
construct a loss function that respects the DCG metric used in ranking.
In fact, \eqref{eq:rankloss_2_bd} can be seen as a generalization of
robust binary classification by applying the transformation on a
\emph{group} of logistic losses instead of a single logistic loss.  In
both robust classification and ranking, the transformation
$\rho_2(\cdot)$ enables models to give up on part of the problem to
achieve better overall performance.

As we discussed in Section~\ref{sec:binary_loss}, however,
transformation of logistic loss using $\rho_2(\cdot)$ results in
Type-II loss function, which is very difficult to optimize.  Hence,
instead of $\rho_2(\cdot)$ we use an alternative transformation
function $\rho_1(\cdot)$, which generates Type-I loss function, to
define the objective function of RoBiRank:
\begin{align}
  \overline{L}_1(\omega) := \sum_{x \in \Xcal} c_x \sum_{y \in \Ycal_x}
  v\rbr{W_{xy}} \cdot \rho_1 \rbr{ \sum_{y' \in \Ycal_x, y' \neq y}
    \sigma_0\rbr{f_\omega(x,y) - f_\omega(x,y')}}.
  \label{eq:rankloss_1_bd}
\end{align}
Since $\rho_1(t) \geq \rho_2(t)$ for every $t > 0$, we have
$\overline{L}_1(\omega) \geq \overline{L}_2(\omega) \geq L_2(\omega)$
for every $\omega$. Note that $\overline{L}_1(\omega)$ is continuous and
twice differentiable.  Therefore, standard gradient-based optimization
techniques can be applied to minimize it. As is standard, a regularizer
on $\omega$ can be added to avoid overfitting; for simplicity, we use
the $\ell_2$-norm in our experiments.

\subsection{Standard Learning to Rank Experiments}
\label{sec:SmallItemSet}

We conducted experiments to check the performance of the objective
function \eqref{eq:rankloss_1_bd} in a standard learning to rank
setting, with a small number of labels to rank. We pitch RoBiRank
against the following algorithms: RankSVM \citep{LeeLin13}, the ranking
algorithm of \citet{LeSmo07} (called LSRank in the sequel), InfNormPush
\citep{Rudin09}, IRPush \citep{Aga11}, and 8 standard ranking algorithms
implemented in
RankLib\footnote{http://sourceforge.net/p/lemur/wiki/RankLib} namely
MART, RankNet, RankBoost, AdaRank, CoordAscent, LambdaMART, ListNet and
RandomForests.
We use three sources of datasets: LETOR 3.0 \citep{ChaCha11} , LETOR
4.0\footnote{http://research.microsoft.com/en-us/um/beijing/projects/letor/letor4dataset.aspx}
and YAHOO LTRC \citep{QinLiuXuLi10}, which are standard benchmarks for
learning to rank algorithms. Table~\ref{tb:descriptives} shows their
summary statistics.  Each dataset consists of five folds; we consider
the first fold, and use the training, validation, and test splits
provided. We train with different values of the regularization
parameter, and select a parameter with the best NDCG value on the
validation dataset.  The performance of the model with this parameter on
the test dataset is reported. 
We used an optimized implementation of the L-BFGS algorithm provided by
the Toolkit for Advanced Optimization
(TAO)\footnote{http://www.mcs.anl.gov/research/projects/tao/index.html}
for estimating the parameter of RoBiRank.  For the other algorithms, we
either implemented them using our framework or used the implementations
provided by the authors. 


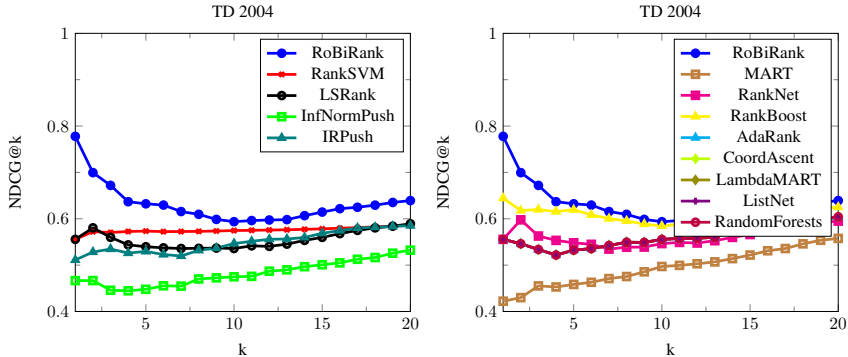
\begin{figure*}[htbp]
  \centering
  \begin{tikzpicture}[scale=0.65]
    \begin{axis}[minor tick num=1,
      title={TD 2004},
      xlabel={k}, ylabel={NDCG@k},
      xmin={1}, xmax={20}, ymin={0.4}, ymax={1.0}]
      
      \addplot[ultra thick, color=blue, mark=*] table [x index=0, y index=1, header=true]
      {./Plots/td2004_all.txt};
      
      \addplot[ultra thick, color=red, mark=x] table [x index=0, y index=2, header=true]
      {./Plots/td2004_all.txt};
      
      \addplot[ultra thick, color=black, mark=o] table [x index=0, y index=3, header=true]
      {./Plots/td2004_all.txt};
      
      \addplot[ultra thick, color=green, mark=square] table [x index=0, y index=4, header=true]
      {./Plots/td2004_all.txt};
      
      \addplot[ultra thick, color=teal, mark=triangle] table [x index=0, y index=5, header=true]
      {./Plots/td2004_all.txt};
      \legend{RoBiRank, RankSVM, LSRank, InfNormPush, IRPush}
    \end{axis}
  \end{tikzpicture}
  \begin{tikzpicture}[scale=0.65]
    \begin{axis}[minor tick num=1,
      title={TD 2004},
      xlabel={k}, ylabel={NDCG@k},
      xmin={1}, xmax={20}, ymin={0.4}, ymax={1.0}]
      
      \addplot[ultra thick, color=blue, mark=*] table [x index=0, y index=1, header=true]
      {./Plots/td2004_all.txt};

      \addplot[ultra thick, color=brown, mark=square] table [x index=0, y index=6, header=true]
      {./Plots/td2004_all.txt};
      
      \addplot[ultra thick, color=magenta, mark=square*] table [x index=0, y index=7, header=true]
      {./Plots/td2004_all.txt};

      \addplot[ultra thick, color=yellow, mark=triangle] table [x index=0, y index=8, header=true]
      {./Plots/td2004_all.txt};

      \addplot[ultra thick, color=cyan, mark=triangle*] table [x index=0, y index=9, header=true]
      {./Plots/td2004_all.txt};

      \addplot[ultra thick, color=lime, mark=diamond] table [x index=0, y index=9, header=true]
      {./Plots/td2004_all.txt};

      \addplot[ultra thick, color=olive, mark=diamond*] table [x index=0, y index=9, header=true]
      {./Plots/td2004_all.txt};

      \addplot[ultra thick, color=violet , mark=+] table [x index=0, y index=9, header=true]
      {./Plots/td2004_all.txt};

      \addplot[ultra thick, color=purple, mark=o] table [x index=0, y index=9, header=true]
      {./Plots/td2004_all.txt};
      \legend{RoBiRank, MART, RankNet, RankBoost, AdaRank, CoordAscent, LambdaMART, ListNet, RandomForests}      
    \end{axis}
  \end{tikzpicture}  
  \caption{Comparison of RoBiRank with a number of competing
    algorithms. Plots are split into two for ease of visualization}
  \label{fig:ndcg_all}
\end{figure*}

We use values of NDCG at different levels of truncation as our
evaluation metric \citep{ManRagSch08}; see Figure~\ref{fig:ndcg_all}.
RoBiRank outperforms its competitors on most of the datasets, however,
due to space constraints we only present plots for the TD 2004 dataset
in the main body of the paper.  Other plots can be found in
Appendix~\ref{sec:AdditPlotsLearn}. The performance of RankSVM seems
insensitive to the level of truncation for NDCG. On the other hand,
RoBiRank, which uses non-convex loss function to concentrate its
performance at the top of the ranked list, performs much better
especially at low truncation levels.  It is also interesting to note
that the NDCG@k curve of LSRank is similar to that of RoBiRank, but
RoBiRank consistently outperforms at each level. RoBiRank dominates
Inf-Push and IR-Push at all levels. When compared to standard
algorithms, Figure~\ref{fig:ndcg_all} (right), again RoBiRank
outperforms especially at the top of the list. 

Overall, RoBiRank outperforms IRPush and InfNormPush on all datasets
except TD 2003 and OHSUMED where IRPush seems to fare better at the top
of the list. Compared to the 8 standard algorithms, again RobiRank
either outperforms or performs comparably to the best algorithm except
on two datasets (TD 2003 and HP 2003), where MART and Random Forests
overtake RobiRank at few values of NDCG. We present a summary of the
NDCG values obtained by each algorithm in Table~\ref{tb:descriptives} in
the appendix. On the MSLR30K dataset, some of the additional algorithms
like InfNormPush and IRPush did not complete within the time period
available; indicated by dashes in the table.



\section{Latent Collaborative Retrieval}


For each context $x$ and an item $y \in \Ycal$, the standard problem
setting of learning to rank requires training data to contain feature
vector $\phi(x,y)$ and score $W_{xy}$ assigned on the $x,y$ pair.  When
the number of contexts $\abr{\Xcal}$ or the number of items
$\abr{\Ycal}$ is large, it might be difficult to define $\phi(x,y)$ and
measure $W_{xy}$ for all $x,y$ pairs.  Therefore, in most learning to
rank problems we define the set of \emph{relevant} items $\Ycal_x
\subset \Ycal$ to be much smaller than $\Ycal$ for each context $x$, and
then collect data only for $\Ycal_x$.  Nonetheless, this may not be
realistic in all situations; in a movie recommender system, for example,
for each user \emph{every} movie is somewhat relevant. 

On the other hand, implicit user feedback data is much more abundant.
For example, a lot of users on Netflix would simply watch movie streams
on the system but do not leave an explicit rating.  By the action of
watching a movie, however, they implicitly express their preference.
Such data consist only of positive feedback, unlike traditional learning
to rank datasets which have score $W_{xy}$ between each context-item
pair $x,y$.  Again, we may not be able to extract feature vectors for
each $x,y$ pair.

In such a situation, we can attempt to learn the score function $f(x,y)$
without a feature vector $\phi(x,y)$ by embedding each context and item
in an Euclidean latent space; specifically, we redefine the score
function to be: $f(x,y) := \inner{U_x}{V_y}$, where $U_x \in \RR^d$ is
the embedding of the context $x$ and $V_y \in \RR^d$ is that of the item
$y$.  Then, we can learn these embeddings by a ranking model.  This
approach was introduced in \citet{WesWanWeiBer12}, and was called
\emph{latent collaborative retrieval}.

Now we specialize RoBiRank model for this task.  Let us define
$\Omega$ to be the set of context-item pairs $(x,y)$ which was
observed in the dataset.  Let $v(W_{xy}) = 1$ if $(x,y) \in \Omega$,
and $0$ otherwise; this is a natural choice since the score
information is not available.  For simplicity, we set $c_x = 1$ for
every $x$.  Now RoBiRank~\eqref{eq:rankloss_1_bd} specializes to:
\begin{align}
  \overline{L}_{1}(U, V) = \sum_{(x,y) \in \Omega}
  \rho_1 \rbr{\sum_{y' \neq y}
    \sigma_0(f(U_x,V_y) - f(U_x, V_{y'}))}.
  \label{eq:latent_obj}
\end{align}
Note that now the summation inside the parenthesis of
\eqref{eq:latent_obj} is over all items $\Ycal$ instead of a smaller set
$\Ycal_x$, therefore we omit specifying the range of $y'$ from now
on. To avoid overfitting, a regularizer is added to
\eqref{eq:latent_obj}; for simplicity we use the Frobenius norm of $U$
and $V$ in our experiments.

\subsection{Stochastic Optimization}
\label{ssec:sto_opt}

When the size of the data $\abr{\Omega}$ or the number of items
$\abr{\Ycal}$ is large, however, methods that require exact evaluation
of the function value and its gradient will become very slow since the
evaluation takes $O\rbr{\abr{\Omega} \cdot \abr{\Ycal} }$ computation.
In this case, stochastic optimization methods are desirable
\citep{BotBou11}; in this subsection, we will develop a stochastic
gradient descent algorithm whose complexity is independent of
$\abr{\Omega}$ and $\abr{\Ycal}$.

For simplicity, let $\theta$ be a concatenation of all parameters
$\cbr{U_x}_{x \in \Xcal}$, $\cbr{V_y}_{y \in \Ycal}$.  The gradient
$\nabla_\theta L_1(U,V)$ of \eqref{eq:latent_obj} is
\begin{align*}
  \sum_{(x,y) \in \Omega}
  \nabla_\theta
  \rho_1 \rbr{\sum_{y' \neq y}
    \sigma_0(f(U_x,V_y) - f(U_x, V_{y'}))}.
\end{align*}
Finding an unbiased estimator of the gradient
whose computation is independent of $\abr{\Omega}$ is not difficult;
if we sample a pair $(x,y)$ uniformly from $\Omega$, then it is easy
to see that the following estimator
\begin{align}
  \abr{\Omega} \cdot 
  \nabla_\theta
  \rho_1 \rbr{\sum_{y' \neq y}
    \sigma_0(f(U_x,V_y) - f(U_x, V_{y'}))}
  \label{eq:simple_sg}
\end{align}
is unbiased.  This still involves a summation over $\Ycal$, however,
so it requires $O(\abr{\Ycal})$ calculation.  Since $\rho_1(\cdot)$ is
a nonlinear function it seems unlikely that an unbiased stochastic
gradient which randomizes over $\Ycal$ can be found; nonetheless, to
achieve convergence guarantees of the stochastic gradient
descent algorithm, unbiasedness of the estimator is necessary
\citep{NemJudLanSha09}.

We attack this problem by \emph{linearizing} the objective function by
parameter expansion.  Note the following property of $\rho_1(\cdot)$
\citep{Bouchard07}: 
\begin{align}
  \rho_1(t) = \log_2(t+1) 
  \leq 
  - \log_2 \xi + \frac{\xi \cdot (t+1) - 1}{\log 2}.
  \label{eq:rho_prop}
\end{align}
This holds for any $\xi > 0$, and the bound is tight when $\xi = \frac
1 {t+1}$.  Now introducing an auxiliary parameter $\xi_{xy}$ for each
$(x,y) \in \Omega$ and applying this bound, we obtain an upper bound
of \eqref{eq:latent_obj} as 
\begin{align}
  L(U,V,\xi) := 
  \label{eq:latent_obj_bd}
  \sum_{(x,y) \in \Omega} - \log_2\xi_{xy} + \frac{\xi_{xy} \rbr{
      \sum_{y' \neq y} \sigma_0(f(U_x,V_y) - f(U_x, V_{y'})) + 1} -
    1}{\log 2}.
\end{align}
Now we propose an iterative algorithm in which, each iteration consists
of $(U,V)$-step and $\xi$-step; in the $(U,V)$-step we minimize
\eqref{eq:latent_obj_bd} in $(U,V)$ and in the $\xi$-step we minimize in
$\xi$.  Pseudo-code can be found in Algorithm~\ref{alg:twostage} in
Appendix~\ref{sec:PseudSeriAlgor}.

\paragraph{$(U,V)$-step} The partial derivative of
\eqref{eq:latent_obj_bd} in terms of $U$ and $V$ can be calculated as:
$\nabla_{U,V} L(U,V,\xi) := \frac{1}{\log 2} \sum_{(x,y) \in \Omega}
\xi_{xy} \rbr{ \sum_{y' \neq y} \nabla_{U,V} \sigma_0(f(U_x,V_y) -
  f(U_x, V_{y'}))}$.  Now it is easy to see that the following
stochastic procedure unbiasedly estimates the above
gradient: 
\begin{itemize*}
\item Sample $(x,y)$ uniformly from $\Omega$
\item Sample $y'$ uniformly from $\Ycal \setminus \cbr{y}$
\item Estimate the gradient by 
  \begin{align}
    \frac{\abr{\Omega} \cdot
      (\abr{\Ycal}-1) \cdot \xi_{xy}}{\log 2} \cdot
    \nabla_{U,V}
    \sigma_0(f(U_x,V_y) - f(U_x, V_{y'})).
    \label{eq:fast_sg}
  \end{align}
\end{itemize*}
Therefore a stochastic gradient descent algorithm based on
\eqref{eq:fast_sg} will converge to a local minimum of the objective
function \eqref{eq:latent_obj_bd} with probability one
\citep{RobMon51}.  Note that the time complexity of calculating
\eqref{eq:fast_sg} is independent of $\abr{\Omega}$ and $\abr{\Ycal}$.
Also, it is a function of only $U_x$ and $V_y$; the gradient is zero
in terms of other variables.

\paragraph{$\xi$-step}  

When $U$ and $V$ are fixed, minimization of $\xi_{xy}$ variable is
independent of each other and a simple analytic solution exists:
$\xi_{xy} = \frac 1 {\sum_{y' \neq y} \sigma_0(f(U_x,V_y) - f(U_x,
  V_{y'})) + 1}$.  This of course requires $O(\abr{\Ycal})$ work.  In
principle, we can avoid summation over $\Ycal$ by taking stochastic
gradient in terms of $\xi_{xy}$ as we did for $U$ and $V$.  However,
since the exact solution is very simple to compute and also because most
of the computation time is spent on $(U,V)$-step rather than $\xi$-step,
we found this update rule to be efficient.

\subsection{Parallelization}
\label{ssec:parallel}

The linearization trick in \eqref{eq:latent_obj_bd} not only enables
us to construct an efficient stochastic gradient algorithm, but also
makes possible to efficiently parallelize the algorithm across
multiple number of machines. Due to lack of space, details are relegated
to Appendix~\ref{sec:DescrParallAlgor}. 

\subsection{Experiments}
\label{sec:LargeItemSet}

In this subsection
we use the
Million Song Dataset (MSD)~\cite{BerEllWhiLam13}, which consists of
1,129,318 users ($\abr{\Xcal}$), 386,133 songs ($\abr{\Ycal}$), and
49,824,519 records ($\abr{\Omega}$) of a user $x$ playing a song $y$ in
the training dataset.  The objective is to predict the songs from the
test dataset that a user is going to listen to\footnote{the original
  data also provides the number of times a song was played by a user,
  but we ignored this in our experiment.}.



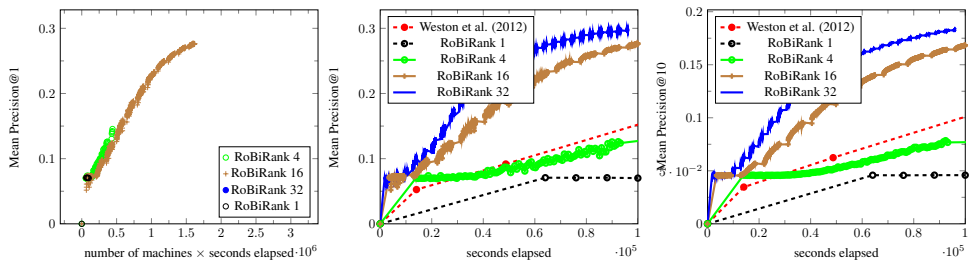
\begin{figure*}
  \centering
  \begin{tikzpicture}[scale=0.5]
    \begin{axis}[minor tick num=1, only marks, 
      xlabel={number of machines $\times$ seconds elapsed},
      ylabel={Mean Precision@1},
      legend style={legend pos=south east}, ymin={0}, mark options={solid} ]
      
      \addplot[color=green, mark=o] table 
      [x expr={\thisrowno{0}*4}, y index=1, header=false, col sep=comma]
      {./Plots/robi_lcr_4.txt};

      \addplot[color=brown, mark=+] table 
      [x expr={\thisrowno{0}*16}, y index=1, header=false, col sep=comma]
      {./Plots/robi_lcr_16.txt};

      \addplot[color=blue, mark=none] table 
      [x expr={\thisrowno{0}*32}, y index=1, header=false, col sep=comma]
      {./Plots/robi_lcr_32.txt};

      \addplot[dashed, color=black, mark=o] table 
      [x expr={\thisrowno{0}*1}, y index=1, header=false, col sep=comma]
      {./Plots/robi_lcr_1.txt};

      \legend{RoBiRank 4, RoBiRank 16, RoBiRank 32, RoBiRank 1}
    \end{axis}
  \end{tikzpicture}
  \begin{tikzpicture}[scale=0.5]
    \begin{axis}[minor tick num=1,
      xlabel={seconds elapsed}, ylabel={Mean Precision@1}, legend pos={north
        west},
      xmin={0}, xmax={100000}, ymin={0}, mark options={solid}]

      \addplot[ultra thick, dashed, color=red, mark=*] table 
      [x index=0, y index=1, header=false, col sep=comma]
      {./Plots/wsabie_time.txt};

      \addplot[ultra thick, dashed, color=black, mark=o] table 
      [x index=0, y index=1, header=false, col sep=comma]
      {./Plots/robi_lcr_1.txt};

      \addplot[ultra thick, color=green, mark=o] table 
      [x index=0, y index=1, header=false, col sep=comma]
      {./Plots/robi_lcr_4.txt};

      \addplot[ultra thick, color=brown, mark=+] table 
      [x index=0, y index=1, header=false, col sep=comma]
      {./Plots/robi_lcr_16.txt};

      \addplot[ultra thick, color=blue, mark=none] table 
      [x index=0, y index=1, header=false, col sep=comma]
      {./Plots/robi_lcr_32.txt};

      \legend{Weston et al. (2012), RoBiRank 1, RoBiRank 4, RoBiRank
        16, RoBiRank 32}
    \end{axis}
  \end{tikzpicture}
  \begin{tikzpicture}[scale=0.5]
    \begin{axis}[minor tick num=1,
      xlabel={seconds elapsed}, ylabel={Mean Precision@10}, legend pos={north
        west},
      xmin={0}, xmax={100000}, ymin={0}, mark options={solid}]

      \addplot[ultra thick, dashed, color=red, mark=*] table 
      [x index=0, y index=2, header=false, col sep=comma]
      {./Plots/wsabie_time.txt};

      \addplot[ultra thick, dashed, color=black, mark=o] table 
      [x index=0, y index=2, header=false, col sep=comma]
      {./Plots/robi_lcr_1.txt};

      \addplot[ultra thick, color=green, mark=o] table 
      [x index=0, y index=2, header=false, col sep=comma]
      {./Plots/robi_lcr_4.txt};

      \addplot[ultra thick, color=brown, mark=+] table 
      [x index=0, y index=2, header=false, col sep=comma]
      {./Plots/robi_lcr_16.txt};

      \addplot[ultra thick, color=blue, mark=none] table 
      [x index=0, y index=2, header=false, col sep=comma]
      {./Plots/robi_lcr_32.txt};

      \legend{Weston et al. (2012), RoBiRank 1, RoBiRank 4, RoBiRank
        16, RoBiRank 32}
    \end{axis}
  \end{tikzpicture}

  \caption{Left:
    the scaling behavior of RoBiRank on Million Song Dataset.  Center, Right:
    Performance comparison of RoBiRank and \citet{WesWanWeiBer12} when
    the same amount of wall-clock time for computation is given.}
  \label{fig:letor_exp}
\end{figure*}

Since explicit ratings are not given, NDCG is not applicable for this
task; we use precision at 1 and 10 \citep{ManRagSch08} as our evaluation
metric. In our first experiment we study the scaling behavior of
RoBiRank as a function of number of machines. RoBiRank $p$ denotes the
parallel version of RoBiRank which is distributed across $p$ machines.
In Figure~\ref{fig:letor_exp} (left) we plot mean Precision@1 as a
function of the number of machines $\times$ the number of seconds
elapsed; this is a proxy for CPU time. If an algorithm linearly scales
across multiple processors, then all lines in the figure should overlap
with each other. As can be seen RoBiRank exhibits near ideal speed up
when going from 4 to 32 machines\footnote{The graph for RoBiRank 1 is
  hard to see because it was run for only 100,000 CPU-seconds.}.

In our next experiment we compare RoBiRank with a state of the art
algorithm from~\citet{WesWanWeiBer12}, which optimizes a similar
objective function \eqref{eq:warp_obj}. We compare how fast the quality
of the solution improves as a function of wall clock time. Since the
authors of \citet{WesWanWeiBer12} do not make available their code, we
implemented their algorithm within our framework using the same data
structures and libraries used by our method. Furthermore, for a fair
comparison, we used the same initialization for $U$ and $V$ and
performed an identical grid-search over the step size parameter for both
algorithms.

It can be seen from Figure~\ref{fig:letor_exp} (center, right) that on a
single machine the algorithm of~\citet{WesWanWeiBer12} is very
competitive and outperforms RoBiRank. The reason for this might be the
introduction of the additional $\xi$ variables in RoBiRank, which slows
down convergence. However, RoBiRank training can be distributed across
processors, while it is not clear how to parallelize the algorithm
of~~\citet{WesWanWeiBer12}. Consequently, RoBiRank 32 which uses 32
machines for its computation can produce a significantly better model
within the same wall clock time window.









\section{Related Work}
\label{sec:RelatedWork}

In terms of modeling, viewing ranking problems as generalization of
binary classification problems is not a new idea; for example, RankSVM
defines the objective function as a sum of hinge losses,
similarly to our basic model \eqref{eq:bound_basic_loss} in
Section~\ref{ssec:basic_model}.  However, it does not directly
optimize the ranking metric such as NDCG; the objective function and
the metric are not immediately related to each other.  In this
respect, our approach is closer to that of \citet{LeSmo07} which
constructs a convex upper bound on the ranking metric and
\citet{ChaDoTeoLeSmo08} which improves the bound by introducing
non-convexity.  The objective function of \citet{ChaDoTeoLeSmo08} is
also motivated by ramp loss, which is used for robust classification;
nonetheless, to our knowledge the direct connection between the
ranking metrics in form \eqref{eq:gen_metric} (DCG, NDCG) and the
robust loss \eqref{robust_losses} is our novel contribution.  Also,
our objective function is designed to specifically bound the ranking
metric, while \citet{ChaDoTeoLeSmo08} proposes a general recipe to
improve existing convex bounds.

Stochastic optimization of the objective function for latent
collaborative retrieval has been also explored in
\citet{WesWanWeiBer12}.  They attempt to minimize
\begin{align}
  \sum_{(x,y) \in \Omega}
  \Phi \rbr{1 + \sum_{y' \neq y}
    I(f(U_x,V_y) - f(U_x, V_{y'}) < 0)},
  \label{eq:warp_obj}
\end{align}
where $\Phi(t) = \sum_{k=1}^t \frac 1 k$.  This is similar to our
objective function \eqref{eq:latent_obj_bd}; $\Phi(\cdot)$ and
$\rho_2(\cdot)$ are asymptotically equivalent.  However, we argue that
our formulation \eqref{eq:latent_obj_bd} has two major advantages.
First, it is a continuous and differentiable function, therefore
gradient-based algorithms such as L-BFGS and stochastic gradient
descent have convergence guarantees.  On the other hand, the objective
function of \citet{WesWanWeiBer12} is not even continuous, since their
formulation is based on a function $\Phi(\cdot)$ that is defined for
only natural numbers.  Also, through the linearization trick in
\eqref{eq:latent_obj_bd} we are able to obtain an unbiased stochastic
gradient, which is necessary for the convergence guarantee, and
to parallelize the algorithm across multiple machines as discussed in
Section~\ref{ssec:parallel}.  It is unclear how these techniques can
be adapted for the objective function of \citet{WesWanWeiBer12}.

\section{Conclusion}
\label{sec:Conclusion}

In this paper, we developed RoBiRank, a novel model on ranking, based
on insights and techniques from the literature of robust binary
classification.  Then, we proposed a scalable and parallelizable
stochastic optimization algorithm that can be applied to the task of
latent collaborative retrieval which large-scale data without feature
vectors and explicit scores have to take care of.  Experimental
results on both learning to rank datasets and latent collaborative
retrieval dataset suggest the advantage of our approach.



As a final note, the experiments in Section~\ref{sec:LargeItemSet} are
arguably unfair towards WSABIE. For instance, one could envisage using
clever engineering tricks to derive a parallel variant of WSABIE (\eg,
by averaging gradients from various machines), which might outperform
RoBiRank on the MSD dataset. While performance on a specific dataset
might be better, we would lose global convergence guarantees. Therefore,
rather than obsess over the performance of a specific algorithm on a
specific dataset, via this paper we hope to draw the attention of the
community to the need for developing principled parallel algorithms for
this important problem.


\bibliographystyle{abbrvnat}
\bibliography{rankbib}

\begin{thebibliography}{24}
\providecommand{\natexlab}[1]{#1}
\providecommand{\url}[1]{\texttt{#1}}
\expandafter\ifx\csname urlstyle\endcsname\relax
  \providecommand{\doi}[1]{doi: #1}\else
  \providecommand{\doi}{doi: \begingroup \urlstyle{rm}\Url}\fi

\bibitem[Agarwal(2011)]{Aga11}
S.~Agarwal.
\newblock The infinite push: A new support vector ranking algorithm that
  directly optimizes accuracy at the absolute top of the list.
\newblock In \emph{SDM}, pages 839--850. SIAM, 2011.

\bibitem[Bartlett et~al.(2006)Bartlett, Jordan, and McAuliffe]{JorBarMcA06}
P.~L. Bartlett, M.~I. Jordan, and J.~D. McAuliffe.
\newblock Convexity, classification, and risk bounds.
\newblock \emph{Journal of the American Statistical Association}, 101\penalty0
  (473):\penalty0 138--156, 2006.

\bibitem[Bertin-Mahieux et~al.(2011)Bertin-Mahieux, Ellis, Whitman, and
  Lamere]{BerEllWhiLam13}
T.~Bertin-Mahieux, D.~P. Ellis, B.~Whitman, and P.~Lamere.
\newblock The million song dataset.
\newblock In \emph{{Proceedings of the 12th International Conference on Music
  Information Retrieval ({ISMIR} 2011)}}, 2011.

\bibitem[Bottou and Bousquet(2011)]{BotBou11}
L.~Bottou and O.~Bousquet.
\newblock The tradeoffs of large-scale learning.
\newblock \emph{Optimization for Machine Learning}, page 351, 2011.

\bibitem[Bouchard(2007)]{Bouchard07}
G.~Bouchard.
\newblock Efficient bounds for the softmax function, applications to inference
  in hybrid models.
\newblock 2007.

\bibitem[Boyd and Vandenberghe(2004)]{BoyVan04}
S.~Boyd and L.~Vandenberghe.
\newblock \emph{Convex Optimization}.
\newblock Cambridge University Press, Cambridge, England, 2004.

\bibitem[Buffoni et~al.(2011)Buffoni, Gallinari, Usunier, and
  Calauz{\`e}nes]{BufGalUsuCal11}
D.~Buffoni, P.~Gallinari, N.~Usunier, and C.~Calauz{\`e}nes.
\newblock Learning scoring functions with order-preserving losses and
  standardized supervision.
\newblock In \emph{Proceedings of the 28th International Conference on Machine
  Learning (ICML-11)}, pages 825--832, 2011.

\bibitem[Chapelle and Chang(2011)]{ChaCha11}
O.~Chapelle and Y.~Chang.
\newblock Yahoo! learning to rank challenge overview.
\newblock \emph{Journal of Machine Learning Research-Proceedings Track},
  14:\penalty0 1--24, 2011.

\bibitem[Chapelle et~al.(2008)Chapelle, Do, Teo, Le, and
  Smola]{ChaDoTeoLeSmo08}
O.~Chapelle, C.~B. Do, C.~H. Teo, Q.~V. Le, and A.~J. Smola.
\newblock Tighter bounds for structured estimation.
\newblock In \emph{Advances in neural information processing systems}, pages
  281--288, 2008.

\bibitem[Ding(2013)]{Ding13}
N.~Ding.
\newblock \emph{Statistical Machine Learning in T-Exponential Family of
  Distributions}.
\newblock PhD thesis, PhD thesis, Purdue University, West Lafayette, Indiana,
  USA, 2013.

\bibitem[Feldman et~al.(2012)Feldman, Guruswami, Raghavendra, and
  Wu]{FelGurRagWu12}
V.~Feldman, V.~Guruswami, P.~Raghavendra, and Y.~Wu.
\newblock Agnostic learning of monomials by halfspaces is hard.
\newblock \emph{SIAM Journal on Computing}, 41\penalty0 (6):\penalty0
  1558--1590, 2012.

\bibitem[Gemulla et~al.(2011)Gemulla, Nijkamp, Haas, and
  Sismanis]{GemNijHaaSis11}
R.~Gemulla, E.~Nijkamp, P.~J. Haas, and Y.~Sismanis.
\newblock Large-scale matrix factorization with distributed stochastic gradient
  descent.
\newblock In \emph{Conference on Knowledge Discovery and Data Mining}, pages
  69--77, 2011.

\bibitem[Huber(1981)]{Huber81}
P.~J. Huber.
\newblock \emph{Robust Statistics}.
\newblock John Wiley and Sons, New York, 1981.

\bibitem[Le and Smola(2007)]{LeSmo07}
Q.~V. Le and A.~J. Smola.
\newblock Direct optimization of ranking measures.
\newblock Technical Report 0704.3359, arXiv, April 2007.
\newblock \url{http://arxiv.org/abs/0704.3359}.

\bibitem[Lee and Lin(2013)]{LeeLin13}
C.-P. Lee and C.-J. Lin.
\newblock Large-scale linear ranksvm.
\newblock \emph{Neural Computation}, 2013.
\newblock To Appear.

\bibitem[Long and Servedio(2010)]{LonSer10}
P.~Long and R.~Servedio.
\newblock Random classification noise defeats all convex potential boosters.
\newblock \emph{Machine Learning Journal}, 78\penalty0 (3):\penalty0 287--304,
  2010.

\bibitem[Manning et~al.(2008)Manning, Raghavan, and Sch{\"u}tze]{ManRagSch08}
C.~D. Manning, P.~Raghavan, and H.~Sch{\"u}tze.
\newblock \emph{Introduction to Information Retrieval}.
\newblock Cambridge University Press, 2008.
\newblock URL \url{http://nlp.stanford.edu/IR-book/}.

\bibitem[Nemirovski et~al.(2009)Nemirovski, Juditsky, Lan, and
  Shapiro]{NemJudLanSha09}
A.~Nemirovski, A.~Juditsky, G.~Lan, and A.~Shapiro.
\newblock Robust stochastic approximation approach to stochastic programming.
\newblock \emph{SIAM Journal on Optimization}, 19\penalty0 (4):\penalty0
  1574--1609, 2009.

\bibitem[Nocedal and Wright(2006)]{NocWri06}
J.~Nocedal and S.~J. Wright.
\newblock \emph{Numerical Optimization}.
\newblock Springer Series in Operations Research. Springer, 2nd edition, 2006.

\bibitem[Qin et~al.(2010)Qin, Liu, Xu, and Li]{QinLiuXuLi10}
T.~Qin, T.-Y. Liu, J.~Xu, and H.~Li.
\newblock Letor: A benchmark collection for research on learning to rank for
  information retrieval.
\newblock \emph{Information Retrieval}, 13\penalty0 (4):\penalty0 346--374,
  2010.

\bibitem[Robbins and Monro(1951)]{RobMon51}
H.~E. Robbins and S.~Monro.
\newblock A stochastic approximation method.
\newblock \emph{Annals of Mathematical Statistics}, 22:\penalty0 400--407,
  1951.

\bibitem[Rudin(2009)]{Rudin09}
C.~Rudin.
\newblock The p-norm push: A simple convex ranking algorithm that concentrates
  at the top of the list.
\newblock \emph{The Journal of Machine Learning Research}, 10:\penalty0
  2233--2271, 2009.

\bibitem[Usunier et~al.(2009)Usunier, Buffoni, and Gallinari]{UsuBufGal09}
N.~Usunier, D.~Buffoni, and P.~Gallinari.
\newblock Ranking with ordered weighted pairwise classification.
\newblock In \emph{Proceedings of the International Conference on Machine
  Learning}, 2009.

\bibitem[Weston et~al.(2012)Weston, Wang, Weiss, and
  Berenzweig]{WesWanWeiBer12}
J.~Weston, C.~Wang, R.~Weiss, and A.~Berenzweig.
\newblock Latent collaborative retrieval.
\newblock \emph{arXiv preprint arXiv:1206.4603}, 2012.

\end{thebibliography}
\clearpage
\appendix

\section{Additional Results for Learning to Rank Experiments}
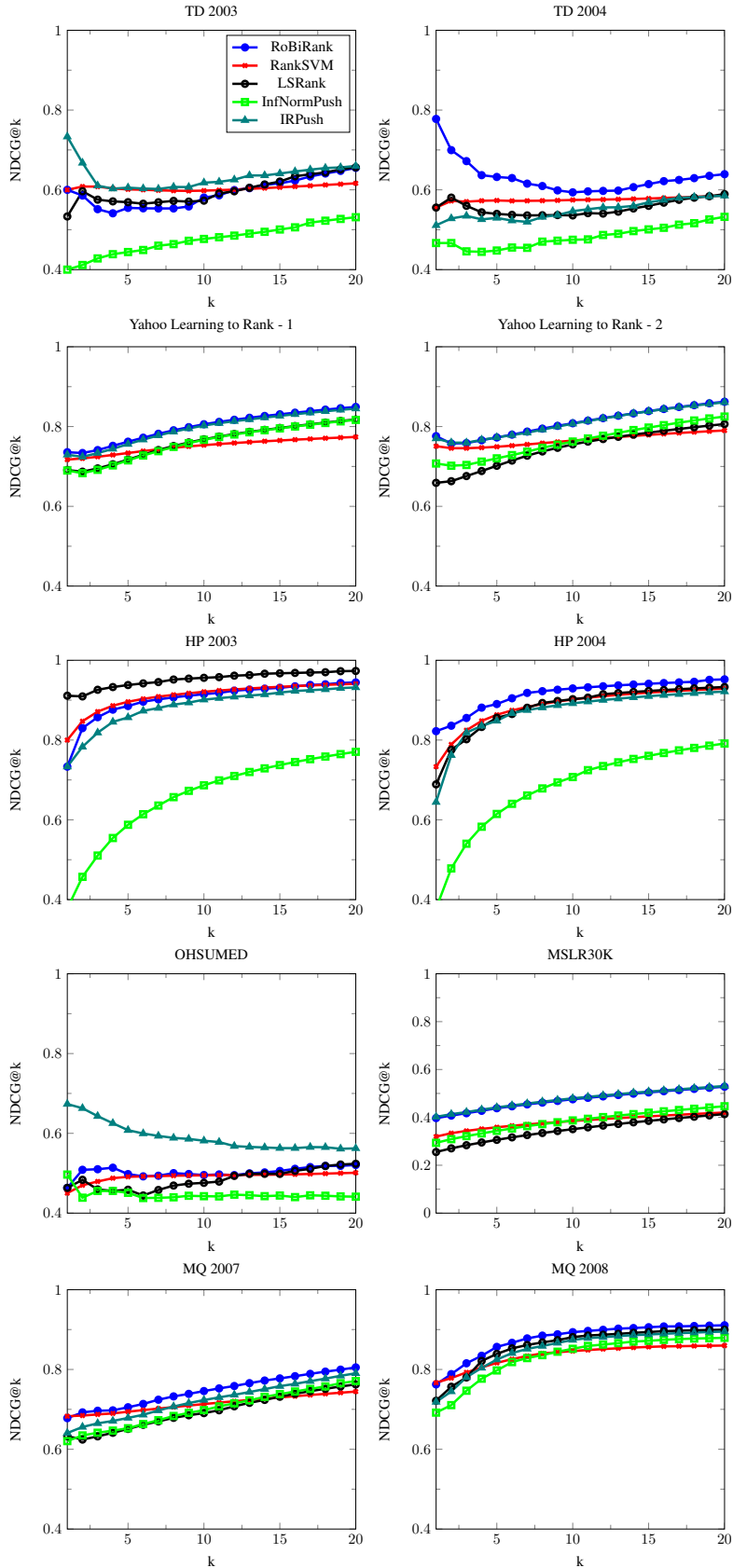
\begin{figure*}[htbp]
  \centering
  \begin{tikzpicture}[scale=0.60]
    \begin{axis}[minor tick num=1,
      title={TD 2003},
      xlabel={k}, ylabel={NDCG@k},
      xmin={1}, xmax={20}, ymin={0.4}, ymax={1.0}]
      
      \addplot[ultra thick, color=blue, mark=*] table [x index=0, y index=1, header=true]
      {./Plots/td2003_all.txt};

      \addplot[ultra thick, color=red, mark=x] table [x index=0, y index=2, header=true]
      {./Plots/td2003_all.txt};

      \addplot[ultra thick, color=black, mark=o] table [x index=0, y index=3, header=true]
      {./Plots/td2003_all.txt};
      
      \addplot[ultra thick, color=green, mark=square] table [x index=0, y index=4, header=true]
      {./Plots/td2003_all.txt};
      
      \addplot[ultra thick, color=teal, mark=triangle] table [x index=0, y index=5, header=true]
      {./Plots/td2003_all.txt};
	  
      \legend{RoBiRank, RankSVM, LSRank, InfNormPush, IRPush}
    \end{axis}
  \end{tikzpicture}
  \begin{tikzpicture}[scale=0.60]
    \begin{axis}[minor tick num=1,
      title={TD 2004},
      xlabel={k}, ylabel={NDCG@k},
      xmin={1}, xmax={20}, ymin={0.4}, ymax={1.0}]
      
      \addplot[ultra thick, color=blue, mark=*] table [x index=0, y index=1, header=true]
      {./Plots/td2004_all.txt};

      \addplot[ultra thick, color=red, mark=x] table [x index=0, y index=2, header=true]
      {./Plots/td2004_all.txt};

      \addplot[ultra thick, color=black, mark=o] table [x index=0, y index=3, header=true]
      {./Plots/td2004_all.txt};
      
      \addplot[ultra thick, color=green, mark=square] table [x index=0, y index=4, header=true]
      {./Plots/td2004_all.txt};
      
      \addplot[ultra thick, color=teal, mark=triangle] table [x index=0, y index=5, header=true]
      {./Plots/td2004_all.txt};
	  
    \end{axis}
  \end{tikzpicture}
  \begin{tikzpicture}[scale=0.60]
    \begin{axis}[minor tick num=1,
      title={Yahoo Learning to Rank - 1},
      xlabel={k}, ylabel={NDCG@k},
      xmin={1}, xmax={20}, ymin={0.4}, ymax={1.0},
      legend pos={south east}]
      
      \addplot[ultra thick, color=blue, mark=*] table [x index=0, y index=1, header=true]
      {./Plots/yahooltrcset1_all.txt};

      \addplot[ultra thick, color=red, mark=x] table [x index=0, y index=2, header=true]
      {./Plots/yahooltrcset1_all.txt};

      \addplot[ultra thick, color=black, mark=o] table [x index=0, y index=3, header=true]
      {./Plots/yahooltrcset1_all.txt};
      
      \addplot[ultra thick, color=green, mark=square] table [x index=0, y index=4, header=true]
      {./Plots/yahooltrcset1_all.txt};
      
      \addplot[ultra thick, color=teal, mark=triangle] table [x index=0, y index=5, header=true]
      {./Plots/yahooltrcset1_all.txt};
	  
    \end{axis}
  \end{tikzpicture}  
  \begin{tikzpicture}[scale=0.60]
    \begin{axis}[minor tick num=1,
      title={Yahoo Learning to Rank - 2},
      xlabel={k}, ylabel={NDCG@k},
      xmin={1}, xmax={20}, ymin={0.4}, ymax={1.0},
      legend pos={south east}]
      
      \addplot[ultra thick, color=blue, mark=*] table [x index=0, y index=1, header=true]
      {./Plots/yahooltrcset2_all.txt};

      \addplot[ultra thick, color=red, mark=x] table [x index=0, y index=2, header=true]
      {./Plots/yahooltrcset2_all.txt};

      \addplot[ultra thick, color=black, mark=o] table [x index=0, y index=3, header=true]
      {./Plots/yahooltrcset2_all.txt};
      
      \addplot[ultra thick, color=green, mark=square] table [x index=0, y index=4, header=true]
      {./Plots/yahooltrcset2_all.txt};
      
      \addplot[ultra thick, color=teal, mark=triangle] table [x index=0, y index=5, header=true]
      {./Plots/yahooltrcset2_all.txt};
	  
    \end{axis}
  \end{tikzpicture}
  \begin{tikzpicture}[scale=0.60]
    \begin{axis}[minor tick num=1,
      title={HP 2003},
      xlabel={k}, ylabel={NDCG@k},
      xmin={1}, xmax={20},
      ymin={0.4}, ymax={1.0},
      legend pos={south east}]
      
      \addplot[ultra thick, color=blue, mark=*] table [x index=0, y index=1, header=true]
      {./Plots/hp2003_all.txt};

      \addplot[ultra thick, color=red, mark=x] table [x index=0, y index=2, header=true]
      {./Plots/hp2003_all.txt};

      \addplot[ultra thick, color=black, mark=o] table [x index=0, y index=3, header=true]
      {./Plots/hp2003_all.txt};
      
      \addplot[ultra thick, color=green, mark=square] table [x index=0, y index=4, header=true]
      {./Plots/hp2003_all.txt};
      
      \addplot[ultra thick, color=teal, mark=triangle] table [x index=0, y index=5, header=true]
      {./Plots/hp2003_all.txt};
	  
    \end{axis}
  \end{tikzpicture}
  \begin{tikzpicture}[scale=0.60]
    \begin{axis}[minor tick num=1,
      title={HP 2004},
      xlabel={k}, ylabel={NDCG@k},
      xmin={1}, xmax={20}, ymin={0.4}, ymax={1.0},
      legend pos={south east}]
      
      \addplot[ultra thick, color=blue, mark=*] table [x index=0, y index=1, header=true]
      {./Plots/hp2004_all.txt};

      \addplot[ultra thick, color=red, mark=x] table [x index=0, y index=2, header=true]
      {./Plots/hp2004_all.txt};

      \addplot[ultra thick, color=black, mark=o] table [x index=0, y index=3, header=true]
      {./Plots/hp2004_all.txt};

      \addplot[ultra thick, color=green, mark=square] table [x index=0, y index=4, header=true]
      {./Plots/hp2004_all.txt};
      
      \addplot[ultra thick, color=teal, mark=triangle] table [x index=0, y index=5, header=true]
      {./Plots/hp2004_all.txt};
      
    \end{axis}
  \end{tikzpicture}
  \begin{tikzpicture}[scale=0.60]
    \begin{axis}[minor tick num=1,
      title={OHSUMED},
      xlabel={k}, ylabel={NDCG@k},
      xmin={1}, xmax={20}, ymin={0.4}, ymax={1.0}]
      
      \addplot[ultra thick, color=blue, mark=*] table [x index=0, y index=1, header=true]
      {./Plots/ohsumed_all.txt};

      \addplot[ultra thick, color=red, mark=x] table [x index=0, y index=2, header=true]
      {./Plots/ohsumed_all.txt};

      \addplot[ultra thick, color=black, mark=o] table [x index=0, y index=3, header=true]
      {./Plots/ohsumed_all.txt};
      
      \addplot[ultra thick, color=green, mark=square] table [x index=0, y index=4, header=true]
      {./Plots/ohsumed_all.txt};
      
      \addplot[ultra thick, color=teal, mark=triangle] table [x index=0, y index=5, header=true]
      {./Plots/ohsumed_all.txt};
	  
    \end{axis}
  \end{tikzpicture}
  \begin{tikzpicture}[scale=0.60]
    \begin{axis}[minor tick num=1,
      title={MSLR30K},
      xlabel={k}, ylabel={NDCG@k},
      xmin={1}, xmax={20},
      ymin={0}, ymax={1.0}]
      
      \addplot[ultra thick, color=blue, mark=*] table [x index=0, y index=1, header=true]
      {./Plots/mslr30K_all.txt};

      \addplot[ultra thick, color=red, mark=x] table [x index=0, y index=2, header=true]
      {./Plots/mslr30K_all.txt};

      \addplot[ultra thick, color=black, mark=o] table [x index=0, y index=3, header=true]
      {./Plots/mslr30K_all.txt};
      
      \addplot[ultra thick, color=green, mark=square] table [x index=0, y index=4, header=true]
      {./Plots/mslr30K_all.txt};
      
      \addplot[ultra thick, color=teal, mark=triangle] table [x index=0, y index=5, header=true]
      {./Plots/mslr30K_all.txt};
      
    \end{axis}
  \end{tikzpicture}
  \begin{tikzpicture}[scale=0.60]
    \begin{axis}[minor tick num=1,
      title={MQ 2007},
      xlabel={k}, ylabel={NDCG@k},
      xmin={1}, xmax={20},
      ymin={0.4}, ymax={1.0}]
      
      \addplot[ultra thick, color=blue, mark=*] table [x index=0, y index=1, header=true]
      {./Plots/mq2007_all.txt};

      \addplot[ultra thick, color=red, mark=x] table [x index=0, y index=2, header=true]
      {./Plots/mq2007_all.txt};

      \addplot[ultra thick, color=black, mark=o] table [x index=0, y index=3, header=true]
      {./Plots/mq2007_all.txt};
      
      \addplot[ultra thick, color=green, mark=square] table [x index=0, y index=4, header=true]
      {./Plots/mq2007_all.txt};
      
      \addplot[ultra thick, color=teal, mark=triangle] table [x index=0, y index=5, header=true]
      {./Plots/mq2007_all.txt};
	  
    \end{axis}
  \end{tikzpicture}
  \begin{tikzpicture}[scale=0.60]
    \begin{axis}[minor tick num=1,
      title={MQ 2008},
      xlabel={k}, ylabel={NDCG@k},
      xmin={1}, xmax={20}, ymin={0.4}, ymax={1.0}]
      
      \addplot[ultra thick, color=blue, mark=*] table [x index=0, y index=1, header=true]
      {./Plots/mq2008_all.txt};

      \addplot[ultra thick, color=red, mark=x] table [x index=0, y index=2, header=true]
      {./Plots/mq2008_all.txt};

      \addplot[ultra thick, color=black, mark=o] table [x index=0, y index=3, header=true]
      {./Plots/mq2008_all.txt};
      
      \addplot[ultra thick, color=green, mark=square] table [x index=0, y index=4, header=true]
      {./Plots/mq2008_all.txt};
      
      \addplot[ultra thick, color=teal, mark=triangle] table [x index=0, y index=5, header=true]
      {./Plots/mq2008_all.txt};
	  
    \end{axis}
  \end{tikzpicture}
  \caption{Comparison of RoBiRank, RankSVM, LSRank \citep{LeSmo07}, Inf-Push and IR-Push }\label{fig:ndcg_all_app}
\end{figure*}

In appendix A, we present results from additional experiments that could not
be accommodated in the main paper due to space constraints. Figure~\ref{fig:ndcg_all_app} shows how RoBiRank
fares against InfNormPush and IRPush on various datasets we used. Figure~\ref{fig:ndcg_all_ranklib_app} shows a similar
comparison against the 8 algorithms present in RankLib. Table~\ref{tb:descriptives} provides descriptive 
statistics of all the datasets we ran our experiments, Overall NDCG values obtained and values of the corresponding
regularization parameters. Overall NDCG values have been omitted for the RankLib algorithms as the library doesn't 
support its calculation directly.
\label{sec:AdditPlotsLearn}

\begin{sidewaystable}
  \centering
  \begin{tabular}{|l|c|c|c|c|c|c|c|c|c|c|c|c|}
    \hline
    \multirow{2}{*}{Name} & \multirow{2}{*}{$\abr{\Xcal}$} & avg. &
    \multicolumn{5}{c|}{Mean NDCG} &     \multicolumn{5}{c|}{Regularization Parameter}
    \\ \cline{4-13}
    & & $\abr{\Ycal_x}$ & RoBiRank & RankSVM & LSRank & InfNormPush & IRPush & RoBiRank &
    RankSVM & LSRank & InfNormPush & IRPush \\
    \hline
    TD 2003 & 50 & 981 & 0.9719 & 0.9219 & 0.9721 & 0.9514 & 0.9685 & $10^{-5}$ &
    $10^{-3}$ & $10^{-1}$ & 1 & $10^{-4}$ \\
    TD 2004 & 75 & 989 & 0.9708 & 0.9084 & 0.9648 & 0.9521 & 0.9601 & $10^{-6}$ &
    $10^{-1}$ & $10^{4}$ & $10^{-2}$ & $10^{-4}$ \\
    Yahoo! 1 & 29,921 & 24 & 0.8921 & 0.7960 & 0.871 & 0.8692 & 0.8892 & $10^{-9}$ &
    $10^{3}$ & $10^{4}$ & 10 & $10^{-9}$ \\
    Yahoo! 2 & 6,330 & 27 & 0.9067 & 0.8126 & 0.8624 & 0.8826 & 0.9068 & $10^{-9}$ &
    $10^{5}$ & $10^{4}$ & 10 & $10^{-7}$ \\
    HP 2003 & 150 & 984 & 0.9960 & 0.9927 & 0.9981 & 0.9832 & 0.9939 & $10^{-3}$ &
    $10^{-1}$ & $10^{-4}$ & 1 & $10^{-2}$ \\
    HP 2004 & 75 & 992 & 0.9967 & 0.9918 & 0.9946 & 0.9863 & 0.9949 & $10^{-4}$ &
    $10^{-1}$ & $10^{2}$ & $10^{-2}$ & $10^{-2}$ \\
    OHSUMED & 106 & 169 & 0.8229 & 0.6626 & 0.8184 & 0.7949 & 0.8417 & $10^{-3}$ &
    $10^{-5}$ & $10^{4}$ & 1 & $10^{-3}$ \\
    MSLR30K & 31,531 & 120 & 0.7812 & 0.5841 & 0.727 & - & - & $1$ &
    $10^{3}$ & $10^{4}$ & - & - \\
    MQ 2007 & 1,692 & 41 & 0.8903 & 0.7950 & 0.8688 & 0.8717 & 0.8810 & $10^{-9}$ &
    $10^{-3}$ & $10^{4}$ & 10 & $10^{-6}$ \\
    MQ 2008 & 784 & 19 & 0.9221 & 0.8703 & 0.9133 & 0.8929 & 0.9052 & $10^{-5}$ &
    $10^{3}$ & $10^{4}$ & 10 & $10^{-5}$ \\
    \hline
  \end{tabular}
  \caption{Descriptive Statistics of Datasets and Experimental Results
    in Section~\ref{sec:SmallItemSet}.}
  \label{tb:descriptives}
\end{sidewaystable}

\begin{table*}
  \centering
  \begin{tabular}{|l|c|c|}
    \hline
    Name & RoBiRank & Identity Loss \\
    \hline
    TD 2003 & 0.9719 & 0.9575 \\
    TD 2004 & 0.9708 & 0.9456 \\
    HP 2003 & 0.9960 & 0.9855 \\
    HP 2004 & 0.9967 & 0.9841 \\ 
    MQ 2007 & 0.8903 & 0.7973 \\ 
    MQ 2008 & 0.9221 & 0.8039 \\ 
    \hline
    MSD & 29\% & 17\% \\ 
    \hline
  \end{tabular}
  \caption{Comparison of RoBiRank against Identity Loss as described in Section~\ref{sec:CompBaselines}. We report overall NDCG for experiments on small-medium datasets, while on the million song dataset (MSD) we report Precision@1.}
  \label{tb:otherbaselines}
\end{table*}

\begin{figure*}[htbp]
  \centering
  \begin{tikzpicture}[scale=0.60]
    \begin{axis}[minor tick num=1,
      title={TD 2003},
      xlabel={k}, ylabel={NDCG@k},
      xmin={1}, xmax={20}, ymin={0.4}, ymax={1.0}]
      
      \addplot[ultra thick, color=blue, mark=*] table [x index=0, y index=1, header=true]
      {./Plots/td2003_all.txt};

      \addplot[ultra thick, color=brown, mark=square] table [x index=0, y index=6, header=true]
      {./Plots/td2003_all.txt};
      
      \addplot[ultra thick, color=magenta, mark=square*] table [x index=0, y index=7, header=true]
      {./Plots/td2003_all.txt};

      \addplot[ultra thick, color=yellow, mark=triangle] table [x index=0, y index=8, header=true]
      {./Plots/td2003_all.txt};

      \addplot[ultra thick, color=cyan, mark=triangle*] table [x index=0, y index=9, header=true]
      {./Plots/td2003_all.txt};

      \addplot[ultra thick, color=lime, mark=diamond] table [x index=0, y index=10, header=true]
      {./Plots/td2003_all.txt};

      \addplot[ultra thick, color=olive, mark=diamond*] table [x index=0, y index=11, header=true]
      {./Plots/td2003_all.txt};

      \addplot[ultra thick, color=violet , mark=+] table [x index=0, y index=12, header=true]
      {./Plots/td2003_all.txt};

      \addplot[ultra thick, color=purple, mark=o] table [x index=0, y index=13, header=true]
      {./Plots/td2003_all.txt};
	  	  	  
      \legend{RoBiRank, MART, RankNet, RankBoost, AdaRank, CoordAscent, LambdaMART, ListNet, RandomForests}
    \end{axis}
  \end{tikzpicture}
  \begin{tikzpicture}[scale=0.60]
    \begin{axis}[minor tick num=1,
      title={TD 2004},
      xlabel={k}, ylabel={NDCG@k},
      xmin={1}, xmax={20}, ymin={0.4}, ymax={1.0}]
      
      \addplot[ultra thick, color=blue, mark=*] table [x index=0, y index=1, header=true]
      {./Plots/td2004_all.txt};

      \addplot[ultra thick, color=brown, mark=square] table [x index=0, y index=6, header=true]
      {./Plots/td2004_all.txt};
      
      \addplot[ultra thick, color=magenta, mark=square*] table [x index=0, y index=7, header=true]
      {./Plots/td2004_all.txt};

      \addplot[ultra thick, color=yellow, mark=triangle] table [x index=0, y index=8, header=true]
      {./Plots/td2004_all.txt};

      \addplot[ultra thick, color=cyan, mark=triangle*] table [x index=0, y index=9, header=true]
      {./Plots/td2004_all.txt};

      \addplot[ultra thick, color=lime, mark=diamond] table [x index=0, y index=10, header=true]
      {./Plots/td2004_all.txt};

      \addplot[ultra thick, color=olive, mark=diamond*] table [x index=0, y index=11, header=true]
      {./Plots/td2004_all.txt};

      \addplot[ultra thick, color=violet , mark=+] table [x index=0, y index=12, header=true]
      {./Plots/td2004_all.txt};

      \addplot[ultra thick, color=purple, mark=o] table [x index=0, y index=13, header=true]
      {./Plots/td2004_all.txt};
      
    \end{axis}
  \end{tikzpicture}
  \begin{tikzpicture}[scale=0.60]
    \begin{axis}[minor tick num=1,
      title={Yahoo Learning to Rank - 1},
      xlabel={k}, ylabel={NDCG@k},
      xmin={1}, xmax={20}, ymin={0.4}, ymax={1.0},
      legend pos={south east}]
      
      \addplot[ultra thick, color=blue, mark=*] table [x index=0, y index=1, header=true]
      {./Plots/yahooltrcset1_all.txt};

      \addplot[ultra thick, color=brown, mark=square] table [x index=0, y index=6, header=true]
      {./Plots/yahooltrcset1_all.txt};
      
      \addplot[ultra thick, color=magenta, mark=square*] table [x index=0, y index=7, header=true]
      {./Plots/yahooltrcset1_all.txt};

      \addplot[ultra thick, color=yellow, mark=triangle] table [x index=0, y index=8, header=true]
      {./Plots/yahooltrcset1_all.txt};

      \addplot[ultra thick, color=cyan, mark=triangle*] table [x index=0, y index=9, header=true]
      {./Plots/yahooltrcset1_all.txt};

      \addplot[ultra thick, color=lime, mark=diamond] table [x index=0, y index=10, header=true]
      {./Plots/yahooltrcset1_all.txt};

      \addplot[ultra thick, color=olive, mark=diamond*] table [x index=0, y index=11, header=true]
      {./Plots/yahooltrcset1_all.txt};

      \addplot[ultra thick, color=violet , mark=+] table [x index=0, y index=12, header=true]
      {./Plots/yahooltrcset1_all.txt};

      \addplot[ultra thick, color=purple, mark=o] table [x index=0, y index=13, header=true]
      {./Plots/yahooltrcset1_all.txt};
      
    \end{axis}
  \end{tikzpicture}  
  \begin{tikzpicture}[scale=0.60]
    \begin{axis}[minor tick num=1,
      title={Yahoo Learning to Rank - 2},
      xlabel={k}, ylabel={NDCG@k},
      xmin={1}, xmax={20}, ymin={0.4}, ymax={1.0},
      legend pos={south east}]
      
      \addplot[ultra thick, color=blue, mark=*] table [x index=0, y index=1, header=true]
      {./Plots/yahooltrcset2_all.txt};

      \addplot[ultra thick, color=brown, mark=square] table [x index=0, y index=6, header=true]
      {./Plots/yahooltrcset2_all.txt};
      
      \addplot[ultra thick, color=magenta, mark=square*] table [x index=0, y index=7, header=true]
      {./Plots/yahooltrcset2_all.txt};

      \addplot[ultra thick, color=yellow, mark=triangle] table [x index=0, y index=8, header=true]
      {./Plots/yahooltrcset2_all.txt};

      \addplot[ultra thick, color=cyan, mark=triangle*] table [x index=0, y index=9, header=true]
      {./Plots/yahooltrcset2_all.txt};

      \addplot[ultra thick, color=lime, mark=diamond] table [x index=0, y index=10, header=true]
      {./Plots/yahooltrcset2_all.txt};

      \addplot[ultra thick, color=olive, mark=diamond*] table [x index=0, y index=11, header=true]
      {./Plots/yahooltrcset2_all.txt};

      \addplot[ultra thick, color=violet , mark=+] table [x index=0, y index=12, header=true]
      {./Plots/yahooltrcset2_all.txt};

      \addplot[ultra thick, color=purple, mark=o] table [x index=0, y index=13, header=true]
      {./Plots/yahooltrcset2_all.txt};
      
    \end{axis}
  \end{tikzpicture}
  \begin{tikzpicture}[scale=0.60]
    \begin{axis}[minor tick num=1,
      title={HP 2003},
      xlabel={k}, ylabel={NDCG@k},
      xmin={1}, xmax={20},
      ymin={0.4}, ymax={1.0},
      legend pos={south east}]
      
      \addplot[ultra thick, color=blue, mark=*] table [x index=0, y index=1, header=true]
      {./Plots/hp2003_all.txt};

     \addplot[ultra thick, color=brown, mark=square] table [x index=0, y index=6, header=true]
      {./Plots/hp2003_all.txt};
      
      \addplot[ultra thick, color=magenta, mark=square*] table [x index=0, y index=7, header=true]
      {./Plots/hp2003_all.txt};

      \addplot[ultra thick, color=yellow, mark=triangle] table [x index=0, y index=8, header=true]
      {./Plots/hp2003_all.txt};

      \addplot[ultra thick, color=cyan, mark=triangle*] table [x index=0, y index=9, header=true]
      {./Plots/hp2003_all.txt};

      \addplot[ultra thick, color=lime, mark=diamond] table [x index=0, y index=10, header=true]
      {./Plots/hp2003_all.txt};

      \addplot[ultra thick, color=olive, mark=diamond*] table [x index=0, y index=11, header=true]
      {./Plots/hp2003_all.txt};

      \addplot[ultra thick, color=violet , mark=+] table [x index=0, y index=12, header=true]
      {./Plots/hp2003_all.txt};

      \addplot[ultra thick, color=purple, mark=o] table [x index=0, y index=13, header=true]
      {./Plots/hp2003_all.txt};
      
    \end{axis}
  \end{tikzpicture}
  \begin{tikzpicture}[scale=0.60]
    \begin{axis}[minor tick num=1,
      title={HP 2004},
      xlabel={k}, ylabel={NDCG@k},
      xmin={1}, xmax={20}, ymin={0.4}, ymax={1.0},
      legend pos={south east}]
      
      \addplot[ultra thick, color=blue, mark=*] table [x index=0, y index=1, header=true]
      {./Plots/hp2004_all.txt};

      \addplot[ultra thick, color=brown, mark=square] table [x index=0, y index=6, header=true]
      {./Plots/hp2004_all.txt};
      
      \addplot[ultra thick, color=magenta, mark=square*] table [x index=0, y index=7, header=true]
      {./Plots/hp2004_all.txt};

      \addplot[ultra thick, color=yellow, mark=triangle] table [x index=0, y index=8, header=true]
      {./Plots/hp2004_all.txt};

      \addplot[ultra thick, color=cyan, mark=triangle*] table [x index=0, y index=9, header=true]
      {./Plots/hp2004_all.txt};

      \addplot[ultra thick, color=lime, mark=diamond] table [x index=0, y index=10, header=true]
      {./Plots/hp2004_all.txt};

      \addplot[ultra thick, color=olive, mark=diamond*] table [x index=0, y index=11, header=true]
      {./Plots/hp2004_all.txt};

      \addplot[ultra thick, color=violet , mark=+] table [x index=0, y index=12, header=true]
      {./Plots/hp2004_all.txt};

      \addplot[ultra thick, color=purple, mark=o] table [x index=0, y index=13, header=true]
      {./Plots/hp2004_all.txt};
      
    \end{axis}
  \end{tikzpicture}
  \begin{tikzpicture}[scale=0.60]
    \begin{axis}[minor tick num=1,
      title={OHSUMED},
      xlabel={k}, ylabel={NDCG@k},
      xmin={1}, xmax={20}, ymin={0.4}, ymax={1.0}]
      
      \addplot[ultra thick, color=blue, mark=*] table [x index=0, y index=1, header=true]
      {./Plots/ohsumed_all.txt};

      \addplot[ultra thick, color=brown, mark=square] table [x index=0, y index=6, header=true]
      {./Plots/ohsumed_all.txt};
      
      \addplot[ultra thick, color=magenta, mark=square*] table [x index=0, y index=7, header=true]
      {./Plots/ohsumed_all.txt};

      \addplot[ultra thick, color=yellow, mark=triangle] table [x index=0, y index=8, header=true]
      {./Plots/ohsumed_all.txt};

      \addplot[ultra thick, color=cyan, mark=triangle*] table [x index=0, y index=9, header=true]
      {./Plots/ohsumed_all.txt};

      \addplot[ultra thick, color=lime, mark=diamond] table [x index=0, y index=10, header=true]
      {./Plots/ohsumed_all.txt};

      \addplot[ultra thick, color=olive, mark=diamond*] table [x index=0, y index=11, header=true]
      {./Plots/ohsumed_all.txt};

      \addplot[ultra thick, color=violet , mark=+] table [x index=0, y index=12, header=true]
      {./Plots/ohsumed_all.txt};

      \addplot[ultra thick, color=purple, mark=o] table [x index=0, y index=13, header=true]
      {./Plots/ohsumed_all.txt};
      
    \end{axis}
  \end{tikzpicture}
      


      
  \begin{tikzpicture}[scale=0.60]
    \begin{axis}[minor tick num=1,
      title={MQ 2007},
      xlabel={k}, ylabel={NDCG@k},
      xmin={1}, xmax={20},
      ymin={0.4}, ymax={1.0}]
      
      \addplot[ultra thick, color=blue, mark=*] table [x index=0, y index=1, header=true]
      {./Plots/mq2007_all.txt};

      \addplot[ultra thick, color=brown, mark=square] table [x index=0, y index=6, header=true]
      {./Plots/mq2007_all.txt};
      
      \addplot[ultra thick, color=magenta, mark=square*] table [x index=0, y index=7, header=true]
      {./Plots/mq2007_all.txt};

      \addplot[ultra thick, color=yellow, mark=triangle] table [x index=0, y index=8, header=true]
      {./Plots/mq2007_all.txt};

      \addplot[ultra thick, color=cyan, mark=triangle*] table [x index=0, y index=9, header=true]
      {./Plots/mq2007_all.txt};

      \addplot[ultra thick, color=lime, mark=diamond] table [x index=0, y index=10, header=true]
      {./Plots/mq2007_all.txt};

      \addplot[ultra thick, color=olive, mark=diamond*] table [x index=0, y index=11, header=true]
      {./Plots/mq2007_all.txt};

      \addplot[ultra thick, color=violet , mark=+] table [x index=0, y index=12, header=true]
      {./Plots/mq2007_all.txt};

      \addplot[ultra thick, color=purple, mark=o] table [x index=0, y index=13, header=true]
      {./Plots/mq2007_all.txt};
      
    \end{axis}
  \end{tikzpicture}
  \begin{tikzpicture}[scale=0.60]
    \begin{axis}[minor tick num=1,
      title={MQ 2008},
      xlabel={k}, ylabel={NDCG@k},
      xmin={1}, xmax={20}, ymin={0.4}, ymax={1.0}]
      
      \addplot[ultra thick, color=blue, mark=*] table [x index=0, y index=1, header=true]
      {./Plots/mq2008_all.txt};

      \addplot[ultra thick, color=brown, mark=square] table [x index=0, y index=6, header=true]
      {./Plots/mq2008_all.txt};
      
      \addplot[ultra thick, color=magenta, mark=square*] table [x index=0, y index=7, header=true]
      {./Plots/mq2008_all.txt};

      \addplot[ultra thick, color=yellow, mark=triangle] table [x index=0, y index=8, header=true]
      {./Plots/mq2008_all.txt};

      \addplot[ultra thick, color=cyan, mark=triangle*] table [x index=0, y index=9, header=true]
      {./Plots/mq2008_all.txt};

      \addplot[ultra thick, color=lime, mark=diamond] table [x index=0, y index=10, header=true]
      {./Plots/mq2008_all.txt};

      \addplot[ultra thick, color=olive, mark=diamond*] table [x index=0, y index=11, header=true]
      {./Plots/mq2008_all.txt};

      \addplot[ultra thick, color=violet , mark=+] table [x index=0, y index=12, header=true]
      {./Plots/mq2008_all.txt};

      \addplot[ultra thick, color=purple, mark=o] table [x index=0, y index=13, header=true]
      {./Plots/mq2008_all.txt};
      
    \end{axis}
  \end{tikzpicture}
  \caption{Comparison of RoBiRank, MART, RankNet, RankBoost, AdaRank, CoordAscent, LambdaMART, ListNet and RandomForests }\label{fig:ndcg_all_ranklib_app}
\end{figure*}
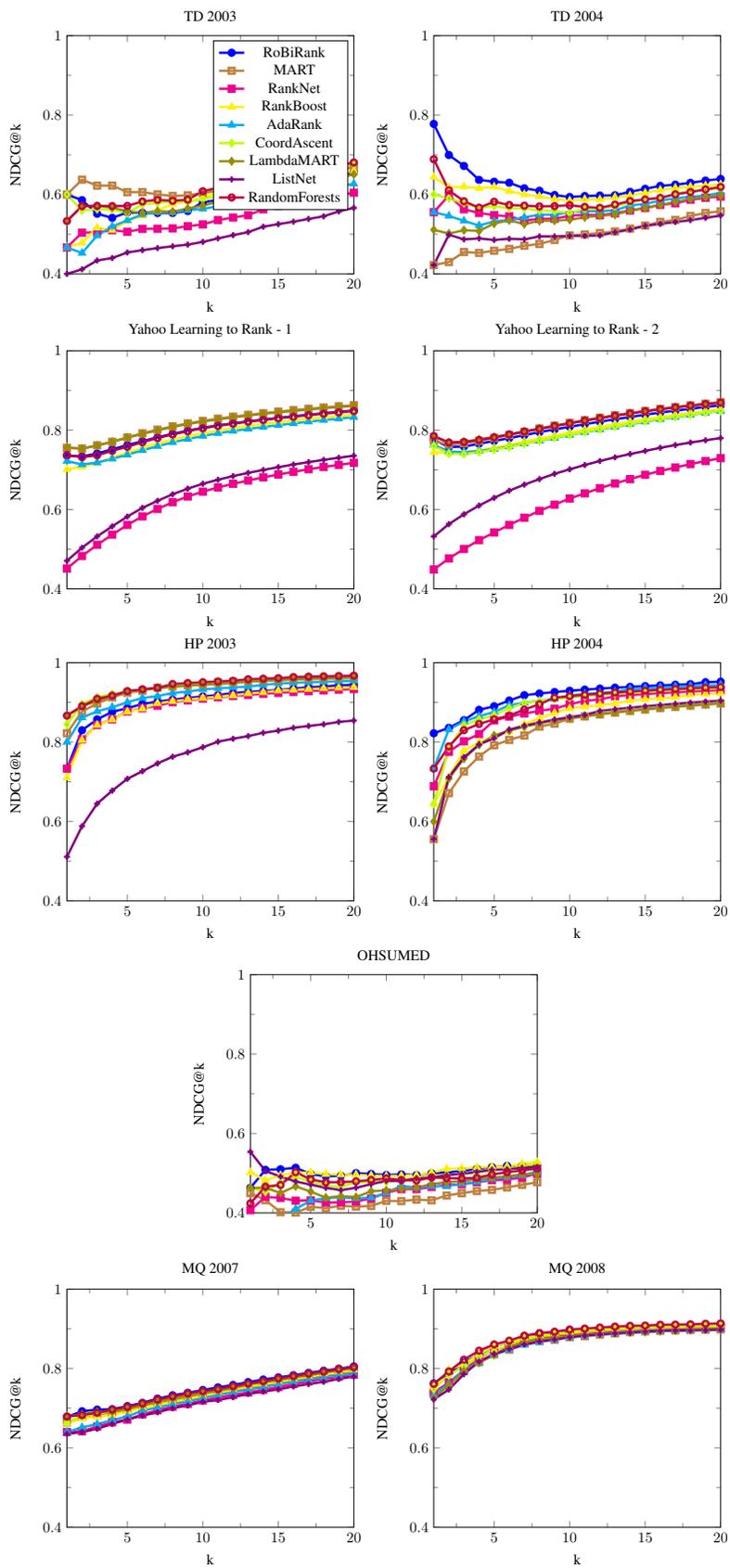

\begin{figure*}[htbp]
  \centering
  \begin{tikzpicture}[scale=0.60]
    \begin{axis}[minor tick num=1,
      title={TD 2003},
      xlabel={k}, ylabel={NDCG@k},
      xmin={1}, xmax={20}, ymin={0.4}, ymax={1.0}]
      
      \addplot[ultra thick, color=blue, mark=*] table [x index=0, y index=1, header=true]
      {./Plots/td2003_all.txt};

      \addplot[ultra thick, color=red, mark=x] table [x index=0, y index=14, header=true]
      {./Plots/td2003_all.txt};

      \legend{RoBiRank, IdentityLoss}
    \end{axis}
  \end{tikzpicture}
  \begin{tikzpicture}[scale=0.60]
    \begin{axis}[minor tick num=1,
      title={TD 2004},
      xlabel={k}, ylabel={NDCG@k},
      xmin={1}, xmax={20}, ymin={0.4}, ymax={1.0}]
      
      \addplot[ultra thick, color=blue, mark=*] table [x index=0, y index=1, header=true]
      {./Plots/td2004_all.txt};

      \addplot[ultra thick, color=red, mark=x] table [x index=0, y index=14, header=true]
      {./Plots/td2004_all.txt};

    \end{axis}
  \end{tikzpicture}
  \begin{tikzpicture}[scale=0.60]
    \begin{axis}[minor tick num=1,
      title={HP 2003},
      xlabel={k}, ylabel={NDCG@k},
      xmin={1}, xmax={20},
      ymin={0.4}, ymax={1.0},
      legend pos={south east}]
      
      \addplot[ultra thick, color=blue, mark=*] table [x index=0, y index=1, header=true]
      {./Plots/hp2003_all.txt};

      \addplot[ultra thick, color=red, mark=x] table [x index=0, y index=14, header=true]
      {./Plots/hp2003_all.txt};

    \end{axis}
  \end{tikzpicture}
  \begin{tikzpicture}[scale=0.60]
    \begin{axis}[minor tick num=1,
      title={HP 2004},
      xlabel={k}, ylabel={NDCG@k},
      xmin={1}, xmax={20}, ymin={0.4}, ymax={1.0},
      legend pos={south east}]
      
      \addplot[ultra thick, color=blue, mark=*] table [x index=0, y index=1, header=true]
      {./Plots/hp2004_all.txt};

      \addplot[ultra thick, color=red, mark=x] table [x index=0, y index=14, header=true]
      {./Plots/hp2004_all.txt};

    \end{axis}
  \end{tikzpicture}
  \begin{tikzpicture}[scale=0.60]
    \begin{axis}[minor tick num=1,
      title={MQ 2007},
      xlabel={k}, ylabel={NDCG@k},
      xmin={1}, xmax={20},
      ymin={0.4}, ymax={1.0}]
      
      \addplot[ultra thick, color=blue, mark=*] table [x index=0, y index=1, header=true]
      {./Plots/mq2007_all.txt};

      \addplot[ultra thick, color=red, mark=x] table [x index=0, y index=14, header=true]
      {./Plots/mq2007_all.txt};
	  
    \end{axis}
  \end{tikzpicture}
  \begin{tikzpicture}[scale=0.60]
    \begin{axis}[minor tick num=1,
      title={MQ 2008},
      xlabel={k}, ylabel={NDCG@k},
      xmin={1}, xmax={20}, ymin={0.4}, ymax={1.0}]
      
      \addplot[ultra thick, color=blue, mark=*] table [x index=0, y index=1, header=true]
      {./Plots/mq2008_all.txt};

      \addplot[ultra thick, color=red, mark=x] table [x index=0, y index=14, header=true]
      {./Plots/mq2008_all.txt};

    \end{axis}
  \end{tikzpicture}
  \caption{Comparison of RoBiRank with other baselines (Identity Loss), see Section~\ref{sec:CompBaselines} }\label{fig:ndcg_otherbaselines}
\end{figure*}
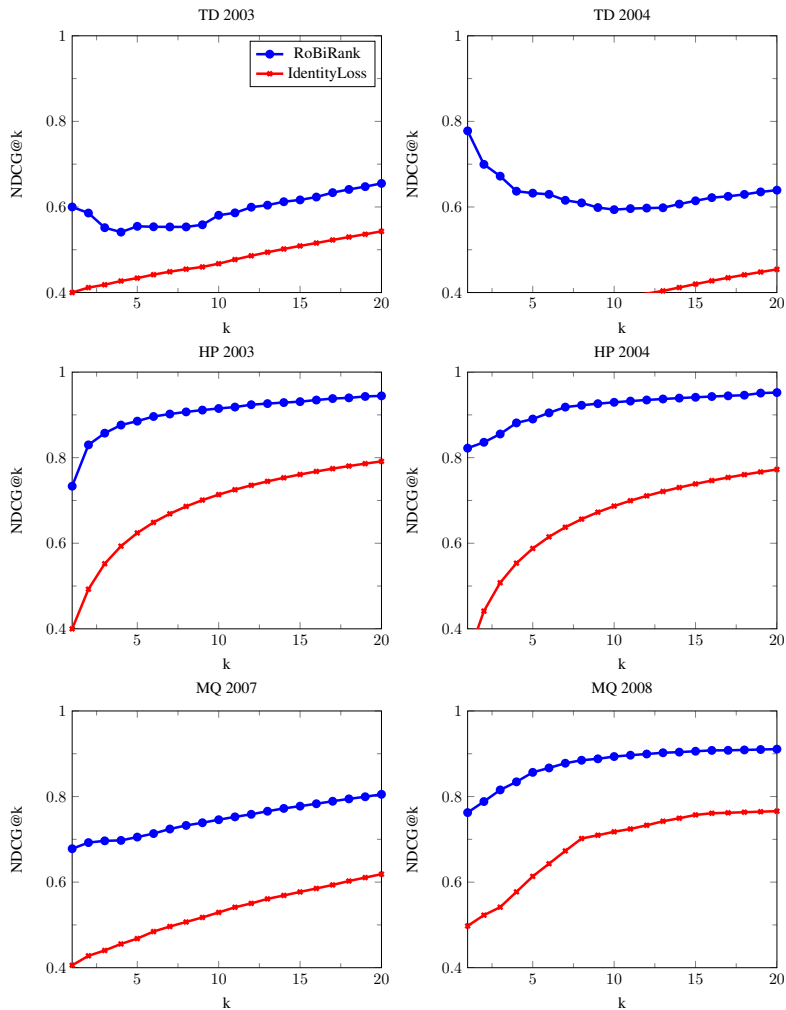

\subsection{Sensitivity to Initialization}
\label{sec:SensInit}

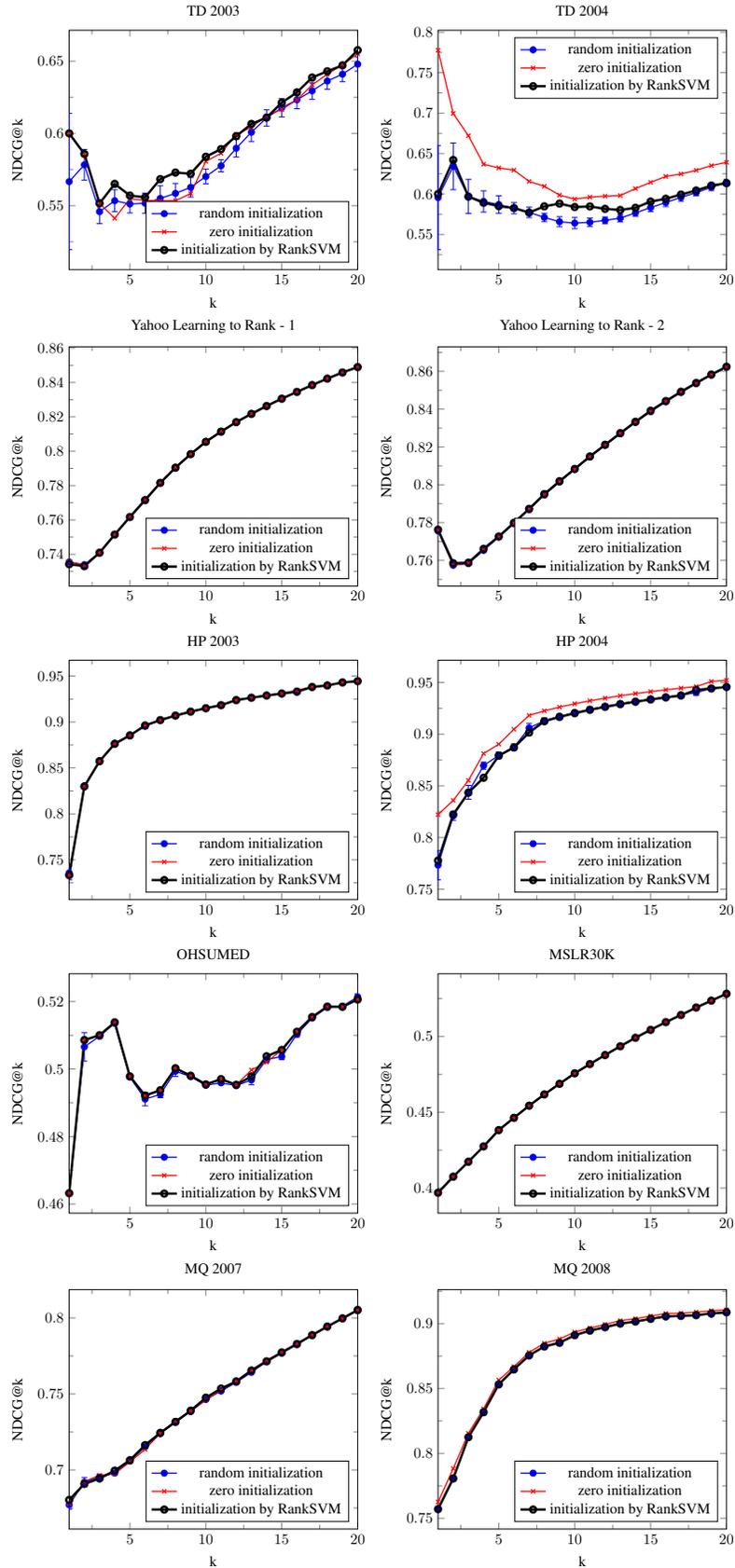
\begin{figure*}
  \centering
  \begin{tikzpicture}[scale=0.60]
    \begin{axis}[minor tick num=1,
      title={TD 2003},
      xlabel={k}, ylabel={NDCG@k},
      xmin={1}, xmax={20}, legend pos={south east}]
      

      \addplot+[
      color=blue, mark=*, thick,
      error bars/.cd,
      y dir=both, y explicit,
      error mark=-]
      table[x=x,y=y,y error=errory]
      {./Plots/td2003errorbar.txt};
      
      \addplot+[thick, mark=x, color=red, mark options=solid] table [x index=0, y index=3, header=true]
      {./Plots/td2003errorbar.txt};

      \addplot+[ultra thick, mark=o, color=black, mark options=solid] table [x index=0, y index=4, header=true]
      {./Plots/td2003errorbar.txt};

      \legend{random initialization, zero initialization, initialization by RankSVM}
    \end{axis}
  \end{tikzpicture}
  \begin{tikzpicture}[scale=0.60]
    \begin{axis}[minor tick num=1,
      title={TD 2004},
      xlabel={k}, ylabel={NDCG@k},
      xmin={1}, xmax={20}, legend pos={north east}]
      

      \addplot+[
      color=blue, mark=*, thick,
      error bars/.cd,
      y dir=both, y explicit,
      error mark=-]
      table[x=x,y=y,y error=errory]
      {./Plots/td2004errorbar.txt};
      
      \addplot+[thick, mark=x, color=red, mark options=solid] table [x index=0, y index=3, header=true]
      {./Plots/td2004errorbar.txt};

      \addplot+[ultra thick, mark=o, color=black, mark options=solid] table [x index=0, y index=4, header=true]
      {./Plots/td2004errorbar.txt};

      \legend{random initialization, zero initialization, initialization by RankSVM}
    \end{axis}
  \end{tikzpicture}
  \begin{tikzpicture}[scale=0.60]
    \begin{axis}[minor tick num=1,
      title={Yahoo Learning to Rank - 1},
      xlabel={k}, ylabel={NDCG@k},
      xmin={1}, xmax={20}, legend pos={south east}]
      

      \addplot+[
      color=blue, mark=*, thick,
      error bars/.cd,
      y dir=both, y explicit,
      error mark=-]
      table[x=x,y=y,y error=errory]
      {./Plots/yahooltrcset1errorbar.txt};
      
      \addplot+[thick, mark=x, color=red, mark options=solid] table [x index=0, y index=3, header=true]
      {./Plots/yahooltrcset1errorbar.txt};

      \addplot+[ultra thick, mark=o, color=black, mark options=solid] table [x index=0, y index=4, header=true]
      {./Plots/yahooltrcset1errorbar.txt};

      \legend{random initialization, zero initialization, initialization by RankSVM}
    \end{axis}
  \end{tikzpicture}
  \begin{tikzpicture}[scale=0.60]
    \begin{axis}[minor tick num=1,
      title={Yahoo Learning to Rank - 2},
      xlabel={k}, ylabel={NDCG@k},
      xmin={1}, xmax={20}, legend pos={south east}]
      

      \addplot+[
      color=blue, mark=*, thick,
      error bars/.cd,
      y dir=both, y explicit,
      error mark=-]
      table[x=x,y=y,y error=errory]
      {./Plots/yahooltrcset2errorbar.txt};
      
      \addplot+[thick, mark=x, color=red, mark options=solid] table [x index=0, y index=3, header=true]
      {./Plots/yahooltrcset2errorbar.txt};

      \addplot+[ultra thick, mark=o, color=black, mark options=solid] table [x index=0, y index=4, header=true]
      {./Plots/yahooltrcset2errorbar.txt};

      \legend{random initialization, zero initialization, initialization by RankSVM}
    \end{axis}
  \end{tikzpicture}
  \begin{tikzpicture}[scale=0.60]
    \begin{axis}[minor tick num=1,
      title={HP 2003},
      xlabel={k}, ylabel={NDCG@k},
      xmin={1}, xmax={20}, legend pos={south east}]
      

      \addplot+[
      color=blue, mark=*, thick,
      error bars/.cd,
      y dir=both, y explicit,
      error mark=-]
      table[x=x,y=y,y error=errory]
      {./Plots/hp2003errorbar.txt};
      
      \addplot+[thick, mark=x, color=red, mark options=solid] table [x index=0, y index=3, header=true]
      {./Plots/hp2003errorbar.txt};

      \addplot+[ultra thick, mark=o, color=black, mark options=solid] table [x index=0, y index=4, header=true]
      {./Plots/hp2003errorbar.txt};

      \legend{random initialization, zero initialization, initialization by RankSVM}
    \end{axis}
  \end{tikzpicture}
  \begin{tikzpicture}[scale=0.60]
    \begin{axis}[minor tick num=1,
      title={HP 2004},
      xlabel={k}, ylabel={NDCG@k},
      xmin={1}, xmax={20}, legend pos={south east}]
      

      \addplot+[
      color=blue, mark=*, thick,
      error bars/.cd,
      y dir=both, y explicit,
      error mark=-]
      table[x=x,y=y,y error=errory]
      {./Plots/hp2004errorbar.txt};
      
      \addplot+[thick, mark=x, color=red, mark options=solid] table [x index=0, y index=3, header=true]
      {./Plots/hp2004errorbar.txt};

      \addplot+[ultra thick, mark=o, color=black, mark options=solid] table [x index=0, y index=4, header=true]
      {./Plots/hp2004errorbar.txt};

      \legend{random initialization, zero initialization, initialization by RankSVM}
    \end{axis}
  \end{tikzpicture}
  \begin{tikzpicture}[scale=0.60]
    \begin{axis}[minor tick num=1,
      title={OHSUMED},
      xlabel={k}, ylabel={NDCG@k},
      xmin={1}, xmax={20}, legend pos={south east}]
      

      \addplot+[
      color=blue, mark=*, thick,
      error bars/.cd,
      y dir=both, y explicit,
      error mark=-]
      table[x=x,y=y,y error=errory]
      {./Plots/ohsumederrorbar.txt};
      
      \addplot+[thick, mark=x, color=red, mark options=solid] table [x index=0, y index=3, header=true]
      {./Plots/ohsumederrorbar.txt};

      \addplot+[ultra thick, mark=o, color=black, mark options=solid] table [x index=0, y index=4, header=true]
      {./Plots/ohsumederrorbar.txt};

      \legend{random initialization, zero initialization, initialization by RankSVM}
    \end{axis}
  \end{tikzpicture}
  \begin{tikzpicture}[scale=0.60]
    \begin{axis}[minor tick num=1,
      title={MSLR30K},
      xlabel={k}, ylabel={NDCG@k},
      xmin={1}, xmax={20}, legend pos={south east}]
      

      \addplot+[
      color=blue, mark=*, thick,
      error bars/.cd,
      y dir=both, y explicit,
      error mark=-]
      table[x=x,y=y,y error=errory]
      {./Plots/mslr30Kerrorbar.txt};
      
      \addplot+[thick, mark=x, color=red, mark options=solid] table [x index=0, y index=3, header=true]
      {./Plots/mslr30Kerrorbar.txt};

      \addplot+[ultra thick, mark=o, color=black, mark options=solid] table [x index=0, y index=4, header=true]
      {./Plots/mslr30Kerrorbar.txt};

      \legend{random initialization, zero initialization, initialization by RankSVM}
    \end{axis}
  \end{tikzpicture}
  \begin{tikzpicture}[scale=0.60]
    \begin{axis}[minor tick num=1,
      title={MQ 2007},
      xlabel={k}, ylabel={NDCG@k},
      xmin={1}, xmax={20}, legend pos={south east}]
      

      \addplot+[
      color=blue, mark=*, thick,
      error bars/.cd,
      y dir=both, y explicit,
      error mark=-]
      table[x=x,y=y,y error=errory]
      {./Plots/mq2007errorbar.txt};
      
      \addplot+[thick, mark=x, color=red, mark options=solid] table [x index=0, y index=3, header=true]
      {./Plots/mq2007errorbar.txt};

      \addplot+[ultra thick, mark=o, color=black, mark options=solid] table [x index=0, y index=4, header=true]
      {./Plots/mq2007errorbar.txt};

      \legend{random initialization, zero initialization, initialization by RankSVM}
    \end{axis}
  \end{tikzpicture}
  \begin{tikzpicture}[scale=0.60]
    \begin{axis}[minor tick num=1,
      title={MQ 2008},
      xlabel={k}, ylabel={NDCG@k},
      xmin={1}, xmax={20}, legend pos={south east}]
      

      \addplot+[
      color=blue, mark=*, thick,
      error bars/.cd,
      y dir=both, y explicit,
      error mark=-]
      table[x=x,y=y,y error=errory]
      {./Plots/mq2008errorbar.txt};
      
      \addplot+[thick, mark=x, color=red, mark options=solid] table [x index=0, y index=3, header=true]
      {./Plots/mq2008errorbar.txt};

      \addplot+[ultra thick, mark=o, color=black, mark options=solid] table [x index=0, y index=4, header=true]
      {./Plots/mq2008errorbar.txt};

      \legend{random initialization, zero initialization, initialization by RankSVM}
    \end{axis}
  \end{tikzpicture}
  \caption{Performance of RoBiRank based on different initialization methods}\label{fig:ranker_all}
\end{figure*}

We also investigated the sensitivity of parameter estimation to the
choice of initial parameter.  We initialized $\omega$ randomly with 10
different seed values.  Blue lines in Figure~\ref{fig:ranker_all} show
mean and standard deviation of NDCG values at different levels of
truncation; as can be seen, even though our objective function is
non-convex, L-BFGS reliably converges to solutions with similar test
performance. This conclusion is in line with the observation
of~\citet{Ding13}.  We also tried two more variants; initialization by
all-zeroes (red line) and the solution of RankSVM (black line).  In most
cases it did not affect the quality of solution, but on TD 2003 and HP
2004 datasets, zero initialization gave slightly better results.

\subsection{Comparison with other baselines}
\label{sec:CompBaselines}

We also compared RoBiRank against other baselines, namely - Identity Loss (obtained by replacing $\rho_1$ by the identity result in the convex loss of \citet{BufGalUsuCal11}). 
We show the results of these experiments on small-medium LETOR datasets and on a large dataset (million song dataset) in Table~\ref{tb:otherbaselines} and Figure~\ref{fig:ndcg_otherbaselines}. As can be seen, RoBiRank comprehensively outperforms these baselines.

\section{Pseudocode of the Serial Algorithm}
\label{sec:PseudSeriAlgor}

\begin{algorithm}
  \begin{algorithmic}
    \STATE {$\eta$: step size}

    \REPEAT

    \STATE {\texttt{// $(U,V)$-step}}
    \REPEAT

    \STATE {Sample $(x,y)$ uniformly from $\Omega$}
    \STATE {Sample $y'$ uniformly from $\Ycal \setminus \cbr{y}$}
    \STATE {$U_x \leftarrow U_x - \eta \cdot \xi_{xy} \cdot 
      \nabla_{U_x} \sigma_0(f(U_x,V_y) - f(U_x, V_{y'}))$}
    \STATE {$V_y \leftarrow V_y - \eta \cdot \xi_{xy} \cdot 
      \nabla_{V_y} \sigma_0(f(U_x,V_y) - f(U_x, V_{y'}))$}
    \UNTIL {convergence in $U,V$}

    \STATE {\texttt{// $\xi$-step}}
    \FOR {$(x,y) \in \Omega$}
    \STATE {$\xi_{xy} \leftarrow \frac 1 
      {\sum_{y' \neq y} \sigma_0(f(U_x,V_y) - f(U_x, V_{y'})) + 1} $}
    \ENDFOR
    \UNTIL {convergence in $U,V$ and $\xi$}
  \end{algorithmic}
  \caption{Serial parameter estimation algorithm for latent
    collaborative retrieval}
  \label{alg:twostage}
\end{algorithm}

\section{Description of Parallel Algorithm}
\label{sec:DescrParallAlgor}

Suppose there are $p$ number of machines.  The set of contexts $\Xcal$
is randomly partitioned into mutually exclusive and exhaustive subsets
$\Xcal^{(1)}, \Xcal^{(2)}, \ldots, \Xcal^{(p)}$ which are of
approximately the same size.  This partitioning is fixed and does not
change over time.  The partition on $\Xcal$ induces partitions on
other variables as follows:
$U^{(q)} := \cbr{ U_x }_{x \in \Xcal^{(q)}}$, 
$\Omega^{(q)} := \cbr{ (x,y) \in \Omega : x \in \Xcal^{(q)}}$,
$\xi^{(q)} := \cbr{ \xi_{xy} }_{(x,y) \in \Omega^{(q)}}$, 
for $1 \leq q \leq p$.

Each machine $q$ stores variables $U^{(q)}$, $\xi^{(q)}$ and
$\Omega^{(q)}$.  Since the partition on $\Xcal$ is fixed, these
variables are local to each machine and are not communicated.  Now we
describe how to parallelize each step of the algorithm: the
pseudo-code can be found in Algorithm~\ref{alg:parallel}.

\begin{algorithm}[!ht]
  \begin{algorithmic}
    \STATE {$\eta$: step size}

    \REPEAT

    \STATE {\texttt{// parallel $(U,V)$-step}}
    \REPEAT

    \STATE {Sample a partition $\cbr{\Ycal^{(1)}, \Ycal^{(2)}, \ldots,
        \Ycal^{(q)}}$}

    \FORALLP {$q \in \cbr{1,2,\ldots,p}$}
    \STATE {Fetch all $V_y \in V^{(q)}$}

    \REPEAT

    \STATE {Sample $(x,y)$ uniformly from\\ \hspace{0.3in} $\cbr{(x,y) \in
        \Omega^{(q)}, y \in \Ycal^{(q)}}$} 
    \STATE {Sample $y'$
      uniformly from $\Ycal^{(q)} \setminus \cbr{y}$} 
    \STATE {$U_x \leftarrow$ \\ \hspace{0.1in} $U_x -
      \eta \cdot \xi_{xy} \cdot \nabla_{U_x} \sigma_0(f(U_x,V_y) -
      f(U_x, V_{y'}))$} 
    \STATE {$V_y \leftarrow$ \\ \hspace{0.1in} $V_y - \eta \cdot
      \xi_{xy} \cdot \nabla_{V_y} \sigma_0(f(U_x,V_y) - f(U_x,
      V_{y'}))$}
    
    \UNTIL {predefined time limit is exceeded}

    \ENDFOR

    \UNTIL {convergence in $U,V$}

    \STATE {\texttt{// parallel $\xi$-step}}

    \FORALLP {$q \in \cbr{1,2,\ldots,p}$}
    \STATE {Fetch all $V_y \in V$}

    \FOR {$(x,y) \in \Omega^{(q)}$}
    \STATE {$\xi_{xy} \leftarrow \frac 1 
      {\sum_{y' \neq y} \sigma_0(f(U_x,V_y) - f(U_x, V_{y'})) + 1} $}
    \ENDFOR

    \ENDFOR

    \UNTIL {convergence in $U,V$ and $\xi$}
  \end{algorithmic}
  \caption{Multi-machine parameter estimation algorithm for latent
    collaborative retrieval}
  \label{alg:parallel}
\end{algorithm}

\paragraph{$(U,V)$-step}

At the start of each $(U,V)$-step, a new partition on $\Ycal$ is sampled
to divide $\Ycal$ into $\Ycal^{(1)}, \Ycal^{(2)}, \ldots, \Ycal^{(p)}$
which are also mutually exclusive, exhaustive and of approximately the
same size.  The difference here is that unlike the partition on
$\Xcal$, a new partition on $\Ycal$ is sampled for every $(U,V)$-step.
Let us define 
$V^{(q)} := \cbr{ V_y }_{y \in \Ycal^{(q)}}$.
After the partition on $\Ycal$ is sampled, each machine $q$ fetches
$V_y$'s in $V^{(q)}$ from where it was previously stored; in the very
first iteration which no previous information exists, each machine
generates and initializes these parameters instead.  Now let us define
$L^{(q)}(U^{(q)},V^{(q)},\xi^{(q)}) :=$
\begin{align*}
  \sum_{(x,y) \in \Omega^{(q)}, y
    \in \Ycal^{(q)}} - \log_2\xi_{xy} + \frac{\xi_{xy} \rbr{ \sum_{y'
        \in \Ycal^{(q)}, y' \neq y} \sigma_0(f(U_x,V_y) - f(U_x,
      V_{y'})) + 1} - 1}{\log 2} .
\end{align*} 
In parallel setting, each machine $q$ runs stochastic gradient descent
on $L^{(q)}(U^{(q)},V^{(q)},\xi^{(q)})$ instead of the original
function $L(U,V,\xi)$.  Since there is no overlap between machines on
the parameters they update and the data they access, every machine can
progress independently of each other.  Although the algorithm takes
only a fraction of data into consideration at a time, this procedure
is also guaranteed to converge to a local optimum of the
\emph{original} function $L(U,V,\xi)$ according to Stratified
Stochastic Gradient Descent (SSGD) scheme of \citet{GemNijHaaSis11}.
The intuition is as follows: if we take expectation over the random
partition on $\Ycal$, we have $  \nabla_{U,V} L(U,V,\xi) = $
\begin{align*}
  q^2 \cdot \EE
  \sbr{ \sum_{1 \leq q \leq p} \nabla_{U,V} L^{(q)}(U^{(q)}, V^{(q)}, \xi^{(q)}) }.
\end{align*}
Therefore, although there is some discrepancy between the function we
take stochastic gradient on and the function we actually aim to
minimize, in the long run the bias will be washed out and the
algorithm will converge to a local optimum of the objective function
$L(U,V,\xi)$.

\paragraph{$\xi$-step}

In this step, all machines synchronize to retrieve every entry of
$V$.  Then, each machine can update $\xi^{(q)}$ independently of each
other.  When the size of $V$ is very large and cannot be fit into the
main memory of a single machine, $V$ can be partitioned as in
$(U,V)$-step and updates can be calculated in a round-robin way.

Note that this parallelization scheme requires each machine to
allocate only $\frac 1 p$-fraction of memory that would be required
for a single-machine execution.  Therefore, in terms of space
complexity the algorithm scales linearly with the number of machines.

\end{document}